\documentclass[10pt,twocolumn,letterpaper]{article}
\usepackage{iccv}     %

\iccvfinalcopy %
\def\sepappendix{0}

\usepackage{times}
\usepackage{epsfig}
\usepackage{graphicx}
\usepackage{amsmath}
\usepackage{amssymb}
\usepackage{booktabs, nicematrix}
\usepackage{makecell}
\usepackage{multirow}
\usepackage{verbatim}
\usepackage{epsfig}
\usepackage{float}
\usepackage{wrapfig}
\usepackage[accsupp]{axessibility}  %
\usepackage[hyphens]{url}
\usepackage{colortbl}
\usepackage{cuted}
\usepackage{bm}
\usepackage{xspace}
\usepackage{multirow}
\usepackage{tabularx}
\usepackage{pifont} %
\usepackage[normalem]{ulem}
\usepackage{arydshln}

\usepackage[pagebackref=true,breaklinks=true,letterpaper=true,colorlinks,bookmarks=false]{hyperref}
\usepackage[numbers,sort,compress]{natbib}

\usepackage[subtle]{savetrees} %

\usepackage{listings}
\definecolor{backcolour}{rgb}{0.95,0.95,0.92}
\lstdefinestyle{mystyle}{
    backgroundcolor=\color{backcolour},  
    keywordstyle=\scriptsize,
    basicstyle=\ttfamily\scriptsize,
    breakatwhitespace=false,
    captionpos=b,                    
    keepspaces=true,                 
    numbers=left,                    
    numbersep=5pt,  
    showspaces=false,                
    showstringspaces=false,
    showtabs=false,                  
    tabsize=4
}

\newcommand{\cam}[1]{\textcolor{black}{#1}}

\newcommand{\methodname}{SINC\xspace}
\newcommand{\methodnamelong}{``SImultaneous actioN Compositions for 3D human motions''\xspace}
\newcommand{\cmark}{\ding{51}}
\newcommand{\xmark}{\ding{55}}

\begin{document}

\title{\methodname{}: Spatial Composition of 3D Human Motions \\ for Simultaneous Action Generation}

\author{Nikos Athanasiou\thanks{Equal contribution}\,\,$^{1}$\quad \;
Mathis Petrovich\textsuperscript{$*$}$^{1,2}$\quad \;
Michael J. Black$^{1}$\quad \;
G{\"u}l Varol$^{2}$\\
$^{1}$Max Planck Institute for Intelligent Systems, T{\"u}bingen, Germany\\
$^{2}$LIGM, {\'E}cole des Ponts, Univ Gustave Eiffel, CNRS, France\\
{ \tt\small
    \href{https://sinc.is.tue.mpg.de}{sinc.is.tue.mpg.de}
}
}

\maketitle
\begin{abstract}
Our goal is to synthesize 3D human motions given textual inputs describing simultaneous actions, for example `waving hand' while `walking' at the same time. We refer to generating such simultaneous movements as performing \textit{spatial compositions}. In contrast to \textit{temporal compositions} that seek to transition from one action to another, spatial compositing requires understanding which body parts are involved in which action, to be able to move them simultaneously. Motivated by the observation that the correspondence between actions and body parts is encoded in powerful language models, we extract this knowledge by prompting GPT-3 with text such as ``what are the body parts involved in the action $<$action name$>$?'', while also providing the parts list and few-shot examples. Given this action-part mapping, we combine body parts from two motions together and establish the first automated method to spatially compose two actions. However, training data with compositional actions is always limited by the combinatorics. Hence, we further create synthetic data with this approach, and use it to train a new state-of-the-art text-to-motion generation model, called \methodname{} (\methodnamelong{}). In our experiments, we find \cam{that training with such GPT-guided synthetic data improves spatial composition generation over baselines.
Our code is publicly available at \href{https://sinc.is.tue.mpg.de/}{sinc.is.tue.mpg.de}.
}
\end{abstract}

\begin{figure}
\centering
    \includegraphics[width=.99\linewidth]{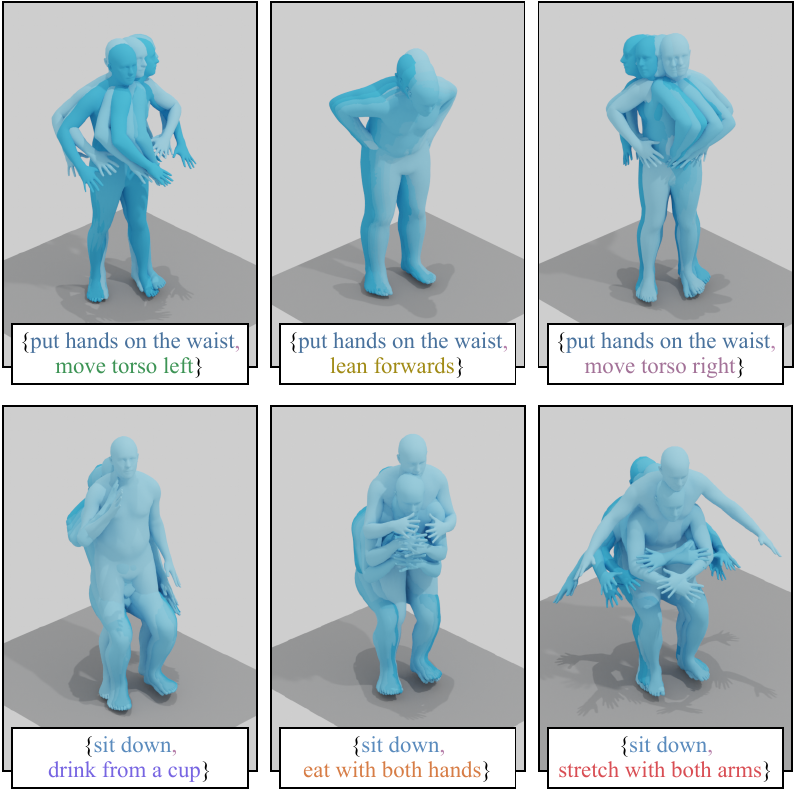}
    \caption{
    \textbf{Goal:} We demonstrate the task of
        \textit{spatial compositions} in human motion synthesis.
        We generate 3D motions for a pair of actions, defined by a pair of textual 
        descriptions.
        Here, we provide six sample input-output illustrations from our model.
        For example, we input the set of actions~\{`put hands on the waist', `move torso left'\}
        and generate one motion that simultaneously performs
        both.
    }
    \label{fig:teaser}
\end{figure}

\section{Introduction}
\label{sec:introduction}

Text-conditioned 3D human motion generation has recently attracted
increasing interest in the research community
\cite{petrovich2022temos,Guo2022GeneratingDA,TEACH:3DV:2022},
where the task is to
input natural language descriptions of actions and to output motion
sequences that semantically correspond to the text. Such controlled
motion synthesis has a variety of applications in fields that rely on motion
capture data, such as special effects, games,
and virtual reality. While there have been promising results
in this direction, \textit{fine-grained} descriptions
remain out of reach.
Consider the scenario in which a movie production needs a particular motion of someone jumping down from a building. 
One may generate an initial motion with one description, and then gradually refine it until  the desired motion is obtained, e.g., ~\{`jumping down', `with arms behind the back',
`while bending the knees'\}. %
State-of-the-art methods \cite{petrovich2022temos, Chen2023}
often fail to produce reasonable motions
when conditioned on fine-grained text describing multiple actions.
In this work, we take a step towards this goal by focusing
on the  \textit{spatial composition} of motions. In other words,
we aim to generate one motion depicting multiple simultaneous actions; see Fig\cam{ure}~\ref{fig:teaser}.
This paves the way for further research on fine-grained human motion generation.

Previous work \cite{lin2018, Ahuja2019Language2PoseNL, Ghosh_2021_ICCV, petrovich2022temos}
initially explored the text-conditioned motion
synthesis problem on the small-scale KIT Motion-Language dataset
\cite{Plappert2016_KIT_ML}.
Recently, work
\cite{Guo2022GeneratingDA,TEACH:3DV:2022} has shifted to the large-scale
motion capture collection AMASS~\cite{AMASS:ICCV:2019}, and
its language labels from BABEL~\cite{BABEL:CVPR:2021} or
HumanML3D~\cite{Guo2022GeneratingDA}.
In particular, similar to this work, TEACH~\cite{TEACH:3DV:2022}
focuses on fine-grained descriptions by addressing temporal
compositionality, that is, generating
a sequence of actions, one \textit{after} the other.
We argue that composition in time is simpler for a model to learn since
the main challenge is to smoothly transition between actions.
This does not necessarily require action-specific knowledge,
and a simple interpolation method such as Slerp~\cite{slerp}
may provide a decent solution. On the other hand, there is no
such trivial solution for compositions in \textit{space}, since one
needs to know action-specific body parts to combine two motions.
If one knows that `waving' involves the hand and `walking' involves
the legs, then compositing the two actions can be performed by
cutting and pasting the hand motion into the walking motion.
This is often done manually in the animation industry.

To automate this process, we observe that pretrained language models
such as GPT-3~\cite{gpt3} encode \cam{knowledge} about which body parts are involved in different actions.
This allows us to first establish a spatial composition baseline
(analogous to the Slerp baseline for temporal compositions); i.e., independently generating actions then combining with heuristics.
Not surprisingly, we find that this is suboptimal.
Instead, we use the synthesized compositions of actions as additional training data for a text-to-motion network. 
This enriched dataset enables our model, called \methodname (\methodnamelong{}), to outperform the baseline.
Our GPT-based approach is similar in spirit to work that incorporates external linguistic knowledge into visual tasks \cite{Yu2017VisualRD,wang2022reconstruction,brooks2022instructpix2pix}.

While BABEL \cite{BABEL:CVPR:2021} and HumanML3D \cite{Guo2022GeneratingDA}
 have relatively large vocabularies of actions,
they contain a limited number of \textit{simultaneous} actions.
A single temporal segment is rarely annotated with multiple texts.
For example, BABEL contains only roughly 2.5K segments with simultaneous actions, %
while it has $\sim$25K segments with only one action.
This highlights the difficulty of obtaining compositional data at scale.
Moreover, for any reasonably large set of actions, it is impractical to collect data
for all possible pairwise, or greater, combinations of actions such that there exists no unseen combination
at test time~\cite{xu2018, Yu2017VisualRD}. 
With existing datasets, it is easy to learn spurious correlations.
For example, if waving is only ever observed by someone standing, a model will learn that waving involves moving the arm with straight legs.
Thus generating waving and sitting would be highly unlikely.
In our work, we address this challenge by artificially creating compositional
data for training using GPT-3. 
By introducing more variety, our generative model is better able to understand what is essential to an action like \cam{`waving'.}

Our method, \methodname, extends the generative text-to-motion model TEMOS~\cite{petrovich2022temos}
such that it
becomes robust to input text describing more than one action, thanks to our synthetic training. We intentionally build on an existing model
to focus the analysis on our proposed synthetic data.
Given a mix of real single actions, real pairs of actions, and synthetic
pairs of actions, we train a probabilistic text-conditioned motion generation
model. 
We introduce several baselines to measure sensitivity to the model design,
as well as to check whether our learned motion decoder outperforms
a simpler compositing technique (i.e., simply using our GPT-guided data creation approach,
\cam{along with a single-action generation model)}. %
We observe limited realism when compositing different body parts
together, and need to incorporate several heuristics, for example
when merging motions whose body parts overlap.
While such synthetic data is imperfect, it helps the model
disentangle the body parts that are relevant for an action and avoid learning spurious correlations.
Moreover, since our motion decoder has also access to real motions, it learns
to generate realistic motions, eliminating the realism problem of the
synthetic composition baseline.

Our contributions are the following:
(i) We establish a new benchmark on the problem of spatial
compositions for 3D human motions, compare a number of baseline models
on this new problem, and introduce a new evaluation metric that is based on a motion encoder that has been trained with text supervision. %
(ii) To address the data scarcity problem, we propose a GPT-guided
synthetic data generation scheme by combining action-relevant body parts
from two motions.
(iii) We provide an extensive set of experiments
on the  BABEL dataset,
including ablations that demonstrate the advantages of our synthetic training, as well as an analysis
quantifying the ability of GPT-3 to assign part labels to actions. 
\cam{Our code is \href{https://sinc.is.tue.mpg.de}{available}} for research purposes.

\section{Related Work}
\label{sec:related}

\noindent\textbf{Human motion generation.}
While motion prediction \cite{Ormoneit2005RepresentingCH, Zhong2022SpatioTemporalGG, Yuan2020DLow, Corona2020ContextAwareHM, Barsoum:2017, Martinez:2017, Salzmann2022MotronMP, Liu2022TowardsRA}, synthesis \cite{Habibie2017rnn, Li2022GANimatorNM} and in-betweening \cite{Harvey2020RobustMI, Tang2022RealtimeCM, Zhou:2020, Kim2022ConditionalMI}
represent the most common motion-generation tasks,
conditional synthesis through other modalities (e.g., text)
has recently received increasing interest. Example conditions include
music \cite{Li2021LearnTD, Moltisanti2022BRACETB}, speech \cite{Habibie2022AMM, Ahn2018Text2ActionGA}, scenes \cite{Starke2019NeuralSM, hassan_samp_2021, zhang2022wanderings, Wang2022TowardsDA}, action \cite{chuan2020action2motion,ACTOR:ICCV:2021}
or text \cite{lin2018,Ahuja2019Language2PoseNL,Ghosh_2021_ICCV,petrovich2022temos,TEACH:3DV:2022,Guo2022GeneratingDA}.
In the following, we focus on work involving 
text-conditioned motion synthesis, which is most closely related to our work.

\begin{figure*}
    \includegraphics[width=\linewidth]{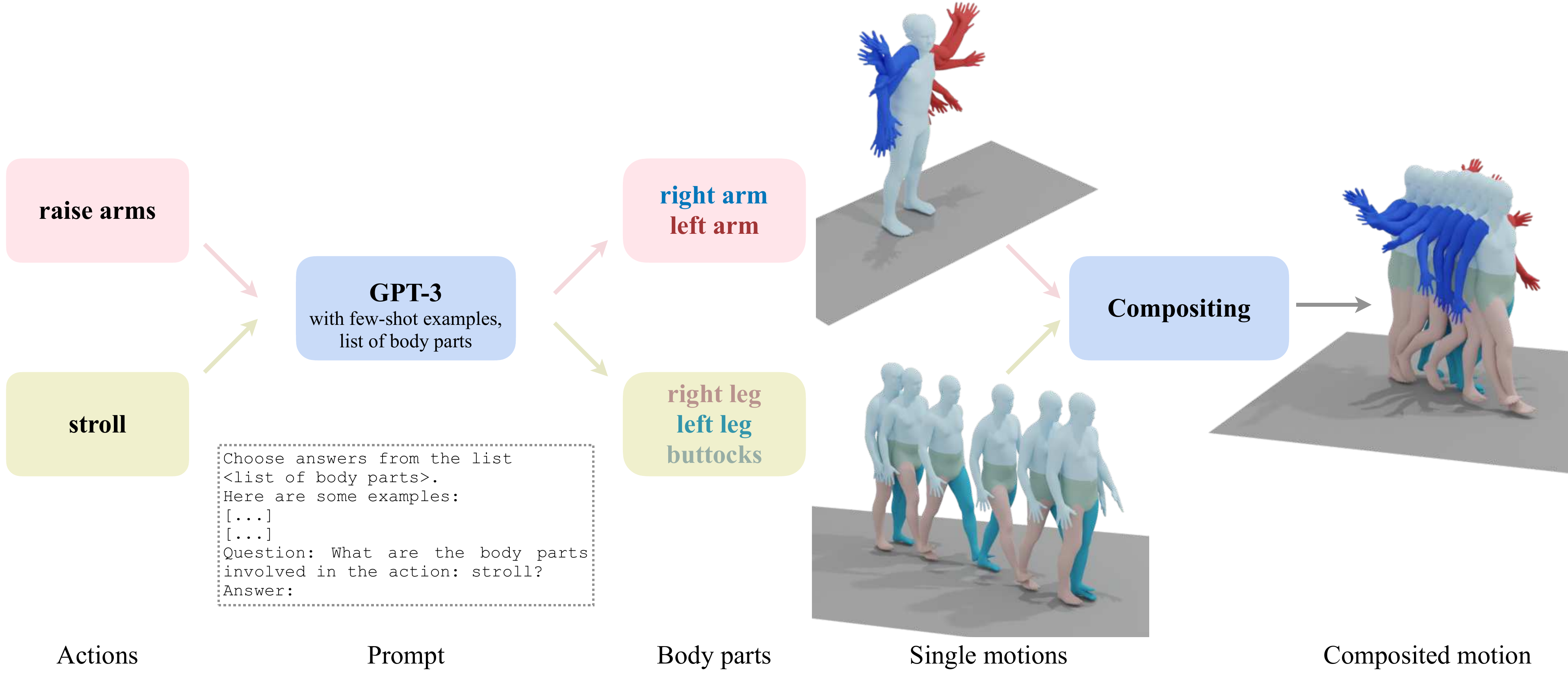}
    \caption{\textbf{GPT-guided synthetic training data creation:} 
        We illustrate our procedure to generate Synth-Pairs. Here, we combine two motion sequences from the training set
        with the corresponding labels `stroll' and `raise arms'. We first prompt GPT-3 with the instructions, few-shot examples containing question-answer pairs, and giving the action of interest in the last question without the answer.
        We minimally post-process the output of GPT-3
        to assign this action
        to a set of body parts.
        The relevant body parts from each motion are then stitched together
        to form a new synthetically composited motion. %
    }
    \label{fig:gpt-synth}
\end{figure*}

\noindent\textbf{3D human motion and natural language.}
Unlike methods that use categorical action labels to control the motion synthesis \cite{chuan2020action2motion,ACTOR:ICCV:2021,poseGPT}, text-conditioned methods
\cite{lin2018,Ahuja2019Language2PoseNL,Ghosh_2021_ICCV,petrovich2022temos,TEACH:3DV:2022,Guo2022GeneratingDA}
seek to input free-form language descriptions that go beyond a closed set of classes.
The KIT-ML dataset~\cite{Plappert2016_KIT_ML} comprises textual annotations
for motion capture data, representing the first benchmark for this task.
More recently, the larger scale AMASS~\cite{AMASS:ICCV:2019} motion capture collection
is labeled with language descriptions by BABEL~\cite{BABEL:CVPR:2021} and
HumanML3D~\cite{Guo2022GeneratingDA}.
A common solution to text-conditioned synthesis is
to design a cross-modal joint space between motions
and language \cite{Ahuja2019Language2PoseNL,Ghosh_2021_ICCV,petrovich2022temos}.
TM2T \cite{Guo_2022_ECCV} introduces a framework
to jointly perform text-to-motion and motion-to-text, integrating a back-translation loss.
In contrast to the deterministic methods of
\cite{Ahuja2019Language2PoseNL,Ghosh_2021_ICCV}, TEMOS~\cite{petrovich2022temos} employs a VAE-based probabilistic approach
(building on ACTOR~\cite{ACTOR:ICCV:2021}) that can generate multiple
motions per textual input,
and establishes the state of the art on the KIT benchmark \cite{Plappert2016_KIT_ML} with
a non-autoregressive architecture.
Following the success of diffusion models \cite{diffusion2015,ddm2020},
very recently, MDM~\cite{Tevet2022HumanMD}, FLAME~\cite{Kim2022FLAMEFL}, MotionDiffuse~\cite{zhang2022motiondiffuse}, and MoFusion \cite{Dabral2022-cq}
demonstrate diffusion-based motion synthesis. %
Recent work~\cite{Chen2023}  shows the potential of latent diffusion
to address the slow inference limitation.
On the other hand, T2M-GPT~\cite{Zhang2023t2mgpt}
obtains competitive performance compared with diffusion using VQ-VAEs.
Our approach is complementary and applicable to existing models for text-to-motion synthesis.
In this work, we adopt TEMOS~\cite{petrovich2022temos}
and retrain it on the data from \cite{BABEL:CVPR:2021} together with our proposed synthetic compositions.

In contrast to previous work, our focus is on the composition of {\em simultaneous} actions.
Prior work  on compositional actions focuses on \textit{temporal} compositions; i.e., inputting a \textit{sequence} of textual descriptions). %
Early influential work \cite{arikan2003siggraph} employs dynamic-programming approaches to compose existing motions from a motion database with action labels.
Recently, Wang et al.~\cite{Wang2022TowardsDA} generate a sequence of actions in 3D scenes
by synthesizing pose anchors that are then placed in the scene and refined by infilling.
TEACH~\cite{TEACH:3DV:2022} extends
TEMOS~\cite{petrovich2022temos} by incorporating an action-level recursive design
that generates the next action conditioned on the past motion.
ActionGPT~\cite{Kalakonda2022-vi} improves this model by retraining it with text augmentations
using language models. 
Concurrently, MultiAct~\cite{Lee2022-multi-act} similarly aims to
produce continuous transitions between generated actions.
In contrast to previous work \cite{TEACH:3DV:2022, Lee2022-multi-act, Kalakonda2022-vi},
we focus on \textit{spatial} compositionality,
inputting text that describes {\em simultaneous actions}.
In this direction,
MotionCLIP~\cite{tevet2022MotionCLIP} and MDM~\cite{Tevet2022HumanMD}
test the compositional capabilities of their methods, but only show preliminary analyses.
The concurrent work of MotionDiffuse~\cite{zhang2022motiondiffuse} injects manually labeled
body-part information and performs noise interpolation to obtain spatial compositionality.

\noindent\textbf{External linguistic knowledge.}
Large language
models have been exploited for many visual tasks
such as
instruction-conditioned image editing~\cite{brooks2022instructpix2pix},
visual relationship detection~\cite{Yu2017VisualRD}, and
human-object reconstruction~\cite{wang2022reconstruction},
among others.
Similar to us, Wang et al.~\cite{wang2022reconstruction}
incorporate GPT by asking what body part is in contact
with a given object, which in turn is used for image-based
3D human-object reconstruction.
On the other hand,
we exploit GPT to extract knowledge about
body parts that are involved in an action.
To the best of our knowledge, we are the first
to systematically model such body part associations 
from textual descriptions.

\noindent\textbf{Training with synthetic data.}
Using synthetic data to train machine
learning models is a standard approach
for solving many visual recognition tasks, such as
3D body pose estimation~\cite{synthetic_cohenor, Patel:CVPR:2021},
2D body part segmentation~\cite{varol17_surreal},
3D hand pose estimation~\cite{hasson19_obman},
video action recognition~\cite{varol21_surreact},
2D body pose estimation \cite{MocapSynth}
pedestrian detection~\cite{SyntheDetect},
and optical flow estimation~\cite{flyingchairs}.
In a similar spirit to us, the recent work of HUMANISE~\cite{Wang2022-jn}
creates a synthetic dataset of
human-scene interactions by combining 4 actions from
BABEL~\cite{BABEL:CVPR:2021} with 3D scenes,
and pairing them with language descriptions.
In this work, we generate
synthetic training data by combining existing 3D motion assets
and language labels to overcome
the data scarcity problem for compositional
learning, helping our method to
avoid learning spurious correlations.

\begin{table*}
    \centering
    \begin{NiceTabular}{lcccccc|c}
    \toprule
        Body part labeling & Global & Torso & Left arm  & Right arm  & Left leg  & Right leg & Mean \\
        \midrule
        \cam{Part velocity magnitude} & 0.72 & 0.68 & 0.60 & 0.55 & 0.58 & 0.67 & 0.65 \\
        \midrule
        \midrule
        \cam{GPT-based} (a) free-form & 0.72 & 0.70 & 0.85 & 0.86 & 0.80 & 0.83 & 0.79 \\
        \cam{GPT-based} (b) choose from list &  0.79 &  0.68 & 0.89 & 0.90 & 0.88 & 0.89 & 0.84 \\
        \cam{GPT-based} (c) choose from list + few-shot examples & \textbf{0.84} & \textbf{0.72} & \textbf{0.89}  & \textbf{0.89} & \textbf{0.89} & \textbf{0.90} & \textbf{0.85} \\ 
    \bottomrule
    \end{NiceTabular}
    \vspace{0.1cm}
    \caption{\textbf{GPT body part labeling performance:}
    We report the part-labeling accuracy of GPT-3, \cam{as well as a simpler baseline
    based on part velocity magnitudes.}
    \cam{For GPT-3, we experiment with}
    various types of prompts on 100 manually annotated actions. (a) Asking which body parts are involved with an action,
    and post-processing free-form language outputs to associate to part labels.
    (b) Asking to choose from a given list of body parts, and
    (c) additionally also providing few-shot examples.
    See Section~\ref{subsec:synth} for more details on these prompts.
    }
    \label{tab:sgpt-labels}
\end{table*}

\noindent\textbf{Compositionality.}
Compositionality has been explored in other areas
of computer vision, such as visual relation detection~\cite{xu2018},
learning object attributes~\cite{Misra2017},
human-object interaction~\cite{Kato2018},
video prediction~\cite{ye2019cvp}, and
video action recognition~\cite{Cruz2020}.
For example,
Shuffle-then-assemble~\cite{xu2018}
explicitly forces their visual relation
detection model to become object-agnostic
to achieve generalization to unseen object pairs.
Similarly, COINS~\cite{Zhao2022-iy} aims
to generate compositions of 
human-scene static interactions, 
where poses that match a text description are generated in a 3D scene.
Here, we focus on action compositionality in space, i.e., simultaneity in time.

\section{Spatial Composition of Motions from Textual Descriptions}
\label{sec:method}

Given a set of action descriptions in the form of text, such as \{``walk in a circle'', ``wave with the right hand''\}, and a desired motion duration $F$,
the goal is to probabilistically generate %
realistic 3D human motions %
such that all the given actions are performed simultaneously
in each generated sequence. 
We refer to this problem as spatial composition.
Note that as a proof of concept, we perform
our experiments mainly with pairs of actions,
but the framework is applicable beyond pairs.

In the following, we first
introduce our framework to
generate synthetic training data by
extracting correspondence between
actions and body parts 
from large language models
(Section~\ref{subsec:synth}). 
Then, we describe our model training with 
synthetically augmented data %
(Section~\ref{subsec:model}),
and finally present implementation details (Section~\ref{subsec:implementation}).

\subsection{GPT-guided synthetic training data creation}
\label{subsec:synth}

As explained in Section~\ref{sec:introduction},
we leverage a large language model, GPT-3 \cite{gpt3},
to automatically assign a given action description
to a set of body parts from a predefined list.
Given such correspondence, we then synthetically combine 
existing motions together to create compositional training data.
This process is illustrated
in \cam{Figure}~\ref{fig:gpt-synth}.

\paragraph{Body part label extraction from GPT-3.}
We process the entire set of motion descriptions in the dataset
to associate each action description to a set of body parts.
We use the Text-Completion tool from  OpenAI's API of GPT-3~\cite{gpt3} to extract the body part
correspondence for a given language description.
Specifically, for each individual action description in the dataset,
we construct a prompt consisting of three parts. (i) We specify the instruction
in the form of ``choose answers from the list $<$list of body parts$>$',
where the list is [`left arm', `right arm', `left leg', `right leg', `torso', `neck', `buttocks', `waist']''.
(ii) We provide few-shot examples as question-answer pairs, where
the question is `What are the body parts involved in the action: $<$action$>$?',
and the answer is the list of manually labeled body parts.
(iii) The last part has the same form as the question, but we do not give the answer.

With this approach, GPT-3 outputs require minimal processing, i.e.,
the responses are words that correspond almost always to the provided list in (i).
We post-process GPT-3's responses by %
removing punctuation, lowercasing, and mapping to
a list of SMPL~\cite{SMPL:2015} body parts that we define separately,
and use in the subsequent steps of our approach to generate synthetic data.
We take a subset of SMPL body parts:
[`left arm', `right arm', `left leg', `right leg', `torso', `global orientation'].
We coarsely define these six different body parts, but
dealing with more fine-grained body parts is certainly possible. %

From the first list, `neck' is mapped to `torso', and [`waist', buttocks']
are mapped to `global orientation'.
This is because, when prompting for free-form outputs without providing a list (i)
or few-shot examples (ii), we qualitatively observe
that GPT-3 refers to changes in global orientation of the body using words such as `waist' or `buttocks'.
Hence, we replace `global orientation' with these two words instead.
GPT-3 also outputs the word `neck' in some cases
even when it is not included in the list, which motivated us to add it
to our list.

To evaluate our choices for the prompt, in Table~\ref{tab:sgpt-labels} 
we measure the contribution
of providing (i) the list, and (ii) few-shot examples in the prompt.
For this, we manually label 100 action descriptions from BABEL.
For each action, we annotate each body part as Yes/No/Sometimes
to mark whether that body part is involved with that action.
Note that we use `Sometimes' for ambiguous cases, where
it is acceptable to include, but not necessarily mandatory.
For example `hands' may or may not be involved in `walking'.
We then check the accuracy of GPT-3 body part labeling,
\cam{by counting Yes/No as 1/0, ignoring optional body parts to not bias our evaluation.}

A prompt asking for a free-form answer
(i.e., ``List the body parts involved in this action: $<$action$>$'') 
complicates the required post-processing as one needs
to handle over-detailed answers such as `deltoids', `triceps',
or different ways of referring to the same body part.
We manually built a lookup table to map from GPT-3 outputs to
SMPL body parts but obtained suboptimal results.
As can be seen from Table~\ref{tab:sgpt-labels},
providing the list (\cam{rows a vs b}) significantly boosts the labeling
accuracy, especially for picking the correct left/right arm/leg, which is further improved by providing
few-shot examples (\cam{row c}).
We provide examples from GPT-3's responses for various prompts
in \if\sepappendix1{Section B}
\else{Section~\ref{sec:supmat:body_parts}}
\fi
of the Appendix.

\vspace{0.2cm}
\noindent\cam{\textit{Could we extract body part labels without GPT-3?}
To test the effectiveness of our GPT-based body part labeling,
we also implement an alternative body-part labeling approach based
on part velocity magnitude. The assumption is that we have action-motion pairs,
and if a body part movement is above a threshold, that part should be involved with the associated action.
Specifically, we compute average positional velocities
across frames for each body part, standardize (subtracting the mean,
dividing by the standard deviation over frames), and determine a threshold (by visual inspection)
to decide if a body part is involved in a given motion.
This heuristic baseline has the disadvantage
that it may suffer from spurious correlations %
(e.g., if we only see waving while walking, we will think that leg
motion is critical to waving).
From the first row of Table~\ref{tab:sgpt-labels},
we observe that the accuracy of this approach is significantly
lower than the GPT-based approaches.
}

\paragraph{Body part composition to create new motions.}
Given a set of labeled motions to combine, and the extracted GPT-3 body
parts involved, we first determine if the actions are compatible;
i.e., whether a valid motion can be composited, based on the descriptions.
For example, the actions [`walking', `kicking with the right leg'] may not
be performed at the same time as they both include the body part `right leg'.
For the synthetic training data, we only create compositions
for valid pairs that are compatible in terms of their body part
involvement, and use \textit{real} motions from the database.
Next, we detail the data creation procedure.

Given two motions A and B, along with the corresponding selected body parts extracted by GPT-3,
we compose these motions into a new one by performing 
the following steps: 
(1) We trim the longer motion to match the length 
of the shorter one; 
(2) We order the motions A and B such that 
motion B always has fewer body parts than motion A; 
(3) If motion B involves at least one leg 
or the global orientation, we also select both legs, 
the global orientation, and translation 
from motion B (otherwise, we obtain these 4 values from motion A);  
(4) The remaining unselected body parts (if any) are taken from motion A;
(5) The composited motion is obtained by combining selected body parts from motion A and B, along with the translation according to step 3.
We perform step 3 to retain plausibility as much as possible,
as the leg motions are highly correlated with changes in global translation and orientation.
This procedure ensures realism and accuracy of the compositions to some extent; but does not provide a guarantee.

Note that we also employ this approach as a baseline in our experiments, where
we combine the motions under these assumptions
using two \textit{generated} motions from a single-action trained model.
In this case, body part incompatibilities may occur
\cam{( ‘walking’ and ‘kicking’ both involve the leg),} and body parts from motion B override
the conflicting parts from motion A
\cam{(see}
\if\sepappendix1{Section~C}
\else{Section~\ref{sec:supmat:synth_data}}
\fi
of the Appendix for further details).

\begin{figure}
    \includegraphics[width=\linewidth]{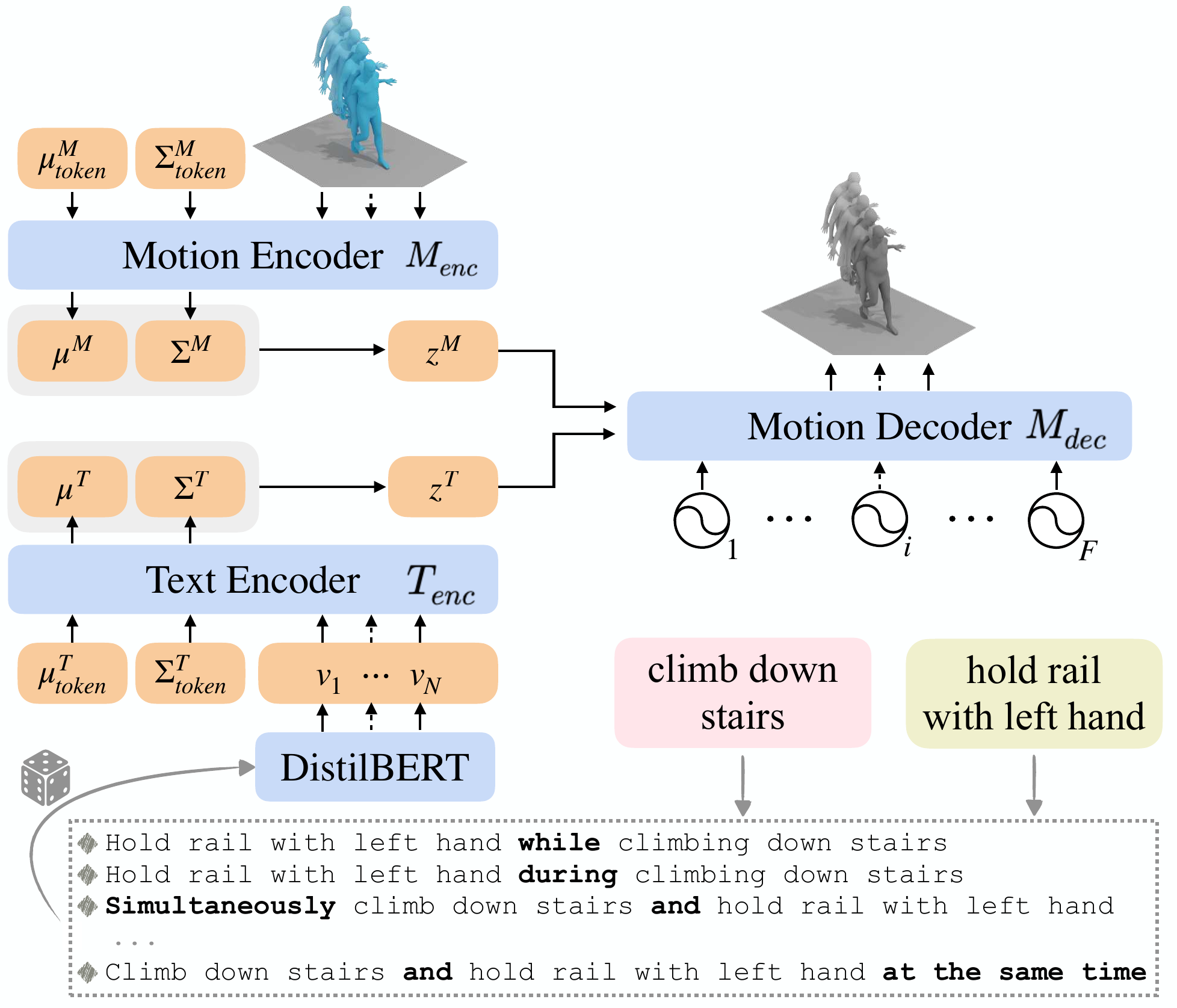}
    \caption{\textbf{Model architecture:}
    We extend TEMOS~\cite{petrovich2022temos} such that it is
    trained with compositional actions.
    We build multiple descriptions given two
    action labels, by adding words such as `while', `during', etc.
    We then randomly sample one version during
    training as input to the text encoder.
    }
    \label{fig:arch}
\end{figure}

\subsection{Learning to generate spatial compositions}
\label{subsec:model}
We employ the recent architecture TEMOS~\cite{petrovich2022temos},
which encodes the text into a distribution via a Transformer encoder
(text encoder $\mathcal{T}_{enc}$),
and produces motions by using a Transformer decoder
(motion decoder $\mathcal{M}_{dec}$). Similar to
Language2Pose~\cite{Ahuja2019Language2PoseNL},
TEMOS contains a motion encoder ($\mathcal{M}_{enc}$) and
encourages a cross-modal joint space between text and motion embeddings.
A simplified overview of the architecture can be seen 
in Figure~\ref{fig:arch}.
At test time, the motion encoder is not used.

The motion encoder takes as input a body motion sequence $B \in \mathbb{R}^{l \times d_f}$ where $d_f$ is the feature dimension and $l$ the maximum motion length and outputs a single latent vector $z^{M}$ and a distribution $\mathcal{N}(\mu^{M} ,\Sigma^{M})$.
Similarly, the text encoder outputs $z^{T}$, which is sampled from the distribution $\mathcal{N}(\mu^{T}$ , $\Sigma^{T})$.
These distribution parameters are obtained by appending two extra learnable tokens in the transformer encoder, and taking their corresponding outputs \cite{ACTOR:ICCV:2021}.
The latent vectors are sampled using the re-parametrization trick~\cite{VAEs}.
The motion decoder then takes as input
(a) the duration encoded by positional encodings $F \in \mathbb{R}^{l \times d}$, where $l$ is the maximum motion length and $d$ the latent dimension, (b) along with either the motion $z^{M}$ or text $z^{T}$ latent vector.

The model is supervised with
the standard normal distribution losses, $\mathcal{L^T_{KL}}{=} \mathcal{KL}(\mathcal{N}(\mu^{T}, \Sigma^{T}), \mathcal{N}(0,I))$ and $\mathcal{L^M_{KL}}{=}\mathcal{KL}(\mathcal{N}(\mu^M, \Sigma^M), \mathcal{N}(0,I)) 
$ for the text and motion distributions, respectively. Moreover, %
$\mathcal{L_{Z}}=\tilde{\mathcal{L}_1}(z^{T}, z^M)$ is used
to force the text latent vectors to be close to the motion latent vector, where $\tilde{\mathcal{L}_1}$ is the smooth L1 loss. Finally, the distributions of different texts and the motion are supervised via $\mathcal{L^{M\|T}_{KL}}{=}\mathcal{KL}(\mathcal{N}(\mu^{T}, \Sigma^{T}),\,\,\, \mathcal{N}(\mu^M, \Sigma^M))$ and its symmetric version $\mathcal{L^{T\|M}_{KL}}$.
The reconstruction losses for the generated motions, $\hat{B}^M$ and $\hat{B}^T$, from both the motion and the text branches, $\mathcal{L_{R}}=
\tilde{\mathcal{L}_1}(B, \hat{B}^T) + \tilde{\mathcal{L}_1}(B, \hat{B}^M)$, are added to the total loss:
\begin{equation}
    \mathcal{L}=\mathcal{L^T_{KL}}+\mathcal{L^M_{KL}}+
    \mathcal{L^{M\|T}_{KL}}+\mathcal{L^{T\|M}_{KL}}+
    \mathcal{L_{R}}+\mathcal{L_{Z}}.
\end{equation}
\cam{
While our experiments use TEMOS~\cite{petrovich2022temos},
our synthetic data strategy is applicable to any text-to-motion
generation model.
We provide further evidence on the benefits of synthetic training
on a diffusion-based approach
(similar to MLD~\cite{Chen2023}) %
in
\if\sepappendix1{Section~A}
\else{Section~\ref{sec:supmat:mld}}
\fi
of the Appendix.
}

\noindent\textbf{Input text format and augmentations.}
Here, we describe how we provide the input to the text encoder.
In case of a single
motion that is described by one action label, we simply
input the original label as in \cite{petrovich2022temos}.
In case of two or more descriptions, which is the focus of this work,
we combine multiple descriptions into a single text.
Specifically, we use several keywords to describe simultaneous actions (e.g., `while', `at the same time', `simultaneously', `during', etc.), and randomly place them in the text description to form an input that imitates a free-form input.
Moreover, we shuffle the order of the labels, and add inflections to verbs such as gerunds when grammatically applicable; e.g., when using `while'. 
Figure~\ref{fig:arch} shows some examples.
Such and input formation allows users to
enter free-form language descriptions at test time,
which is a natural interface for humans.
During training, we pick a random text augmentation,
and at test time, we evaluate all the models using
the conjunction word `while'.
In \if\sepappendix1{Section~E.1}
\else{Section~\ref{subsec:supmat:conjunction}}
\fi
of the Appendix, we provide results with more conjunction words 
\cam{both seen and unseen during training.}

\begin{table*}
    \centering
    \resizebox{\linewidth}{!}{
    \begin{NiceTabular}{l|cc|c|rrrr|rrrr}
        \toprule
        \multirow{2}{*}{\textbf{Model}} & \multicolumn{2}{c}{Tr.~Data} & TEMOS $\uparrow$ & \multicolumn{4}{c|}{Average Positional Error $\downarrow$} & \multicolumn{4}{c}{Average Variance Error $\downarrow$} \\
         & Real-P & Real-S & score & \small{root joint} &  \small{global traj.}  & \small{mean local} & \small{mean global} & \small{root joint} & \small{global traj.} & \small{mean local} & \small{mean global}  \\
        \midrule 
        \textbf{Single-action} & \xmark  & \cmark & 0.601 & 0.592 & 0.551 & 0.286 & 0.712 & 0.076 & 0.075 & 0.013 & 0.083 \\
        \textbf{Single-action GPT-compositing}  & \xmark  & \cmark & 0.618 & 0.546 & 0.507 & 0.282 & 0.666 & 0.076 & 0.075 & 0.013 & 0.082 \\
        \midrule
        \midrule
        \textbf{SINC-STE} & \cmark & \xmark  &  0.614  & 0.636 & 0.615 & 0.275 & 0.743 & 0.082 & 0.081 & 0.014 & 0.090 \\
        \textbf{SINC} & \cmark & \xmark  &  0.631 & 0.703 & 0.682 & 0.269 & 0.815 & 0.107 & 0.106 & 0.013 & 0.114 \\        
        \textbf{SINC}  & \cmark & \cmark &  \textbf{0.640}  & 0.601 & 0.573 & 0.268 & 0.724 & 0.093 & 0.092 & 0.012 & 0.100 \\
        \bottomrule 
    \end{NiceTabular}
    }
\vspace{0.1cm}
    \caption{\textbf{Baseline comparison:} 
    We train only with Real-Pairs of the BABEL dataset and report performance when compositing naively or with GPT-3 annotations. Furthermore, we ablate the model design for handling multiple textual inputs when extending TEMOS~\cite{petrovich2022temos}. We observe better performance at handling action pairs with a single text encoder (SINC) that takes as input the two text labels as a single free-form description with various augmentations, as described in Section~\ref{subsec:model}, compared to separate text encodings of the labels (SINC-STE). Moreover, we report the performance of SINC when adding Real-Singles, as well.
    }
    \label{tab:archi_design}
\end{table*}

\subsection{Implementation details}
\label{subsec:implementation}

We define a 3D human motion 
as a sequence of human poses using the SMPL body model~\cite{SMPL:2015}. %
As in TEMOS~\cite{petrovich2022temos, Holden:2016}, we represent the motion using
the 6D rotations~\cite{zhou2019continuity} for body joints and the 2D-projection of the $x,y$ trajectory along with the $z$ translation. This results in $d_f=135$ for each body pose in each motion sequence. All the motions are canonicalized to face the same forward direction and are standardized.

The input text is encoded with DistilBERT~\cite{distilbert_sanh} 
(whose parameters are frozen), followed by a learnable linear projection.
The latent dimension is fixed to $d=256$. We use $6$ layers and heads in the transformers with a linear projection of size $1024$. We set the batch size to $64$ and the learning rate to $3\cdot10^{-4}$ for all our experiments.

Our model is applicable to arbitrary numbers of actions for a given motion. 
Therefore, we jointly train on single actions, and multiple actions.
Single actions are from real data.
Multiple actions can be (i) from synthetic pairs 
that are randomly generated `on the fly'
or (ii) from real data
where most such motions have two labels, but we
also include those with more than two; see 
the supplementary video \cam{on our project page} for more details.
For each sequence in a mini-batch, if it is a real single action,
with probability $p$, we combine it randomly with another compatible action.

\section{Experiments}
\label{sec:experiments}
We present data and evaluation metrics (Section~\ref{subsec:data}),
followed by the baselines we introduce (Section~\ref{subsec:baselines}).
We report quantitative experimental results with ablations (Sections~\ref{subsec:ablations}~and~\ref{subsec:experiments_data}).
We conclude with a qualitative analysis (Section~\ref{subsec:qualitative})
and a discussion of limitations (Section~\ref{subsec:limitations}).

\begin{table*}
\centering

\resizebox{\linewidth}{!}{
\begin{NiceTabular}{l|ccc|c|rrrr|rrrr}
    \toprule
    \multirow{2}{*}{\textbf{Synthetic data}} & \multicolumn{3}{c}{Training Data} & TEMOS $\uparrow$ & \multicolumn{4}{c|}{Average Positional Error $\downarrow$} & \multicolumn{4}{c}{Average Variance Error $\downarrow$} \\
     & Real-P & Real-S\,\% & Synth-P\,\% & score & \small{root joint} &  \small{global traj.} & \small{mean local} & \small{mean global} & \small{root joint} & \small{global traj.} & \small{mean local} & \small{mean global}  \\
    \midrule
    \textbf{N/A} & \cmark & 0 &  0  &  0.631 & 0.703 & 0.682 & 0.269 & 0.815 & 0.107 & 0.106 & 0.013 & 0.114 \\
    & \cmark & 100 & 0 & 0.640  & 0.601 & 0.573 & 0.268 & 0.724 & 0.093 & 0.092 & 0.012 & 0.100 \\
    \midrule
    \textbf{Random composition} & \xmark & 0 & 100 & 0.539 & 0.489 & 0.434 & 0.291 & 0.595 & 0.075 & 0.074 & 0.012 & 0.082 \\
    & \xmark & 50 & 50 & 0.540 & 0.587 & 0.535 & 0.288 & 0.687 & 0.077 & 0.076 & 0.012 & 0.083 \\
    & \cmark & 0 & 100 & 0.619 & 0.485 & 0.438 & 0.272 & 0.602 & 0.074 & 0.073 & 0.011 & 0.081\\
     & \cmark & 50 & 50 & 0.617 & 0.454 & 0.394 & 0.272 & 0.560 & 0.069 & 0.068 & 0.011 & 0.075 \\
    \midrule
    \textbf{GPT composition} & \xmark & 0 & 100 & 0.618 & 0.478 & 0.451 & 0.265 & 0.610 & 0.063 & 0.062 & 0.012 & 0.070\\
    & \xmark & 50 & 50 & 0.541 & 0.646 & 0.598 & 0.290 & 0.747 & 0.078 & 0.077 & 0.012 & 0.085\\
    & \cmark & 0 & 100 & 0.642 & 0.553 & 0.527 & 0.266 & 0.671 & 0.061 & 0.060 & 0.011 & 0.068 \\
    & \cmark & 50 & 50 & \textbf{0.644} & 0.481 & 0.452 & 0.261 & 0.605 & 0.064 & 0.062 & 0.011 & 0.070 \\
    \bottomrule 
    \end{NiceTabular}
    }
	\vspace{0.1cm}
    \caption{\textbf{Contribution of the synthetic data:} 
    We report performance when including
    two types of synthetic data created by body part combination, either determined by GPT or randomly.
    We further experiment (i) with different percentages of sampling ratios between the Real-Singles and Synth-Pairs,
    and (ii) with the inclusion of Real-Pairs.
    }
    \label{tab:synthetic}
\end{table*}

\subsection{Data and evaluation metrics}
\label{subsec:data}

We use the \textbf{BABEL} dataset~\cite{BABEL:CVPR:2021}, to exploit its unique potential to study simultaneous actions.
Some BABEL motions come with multiple %
language descriptions where annotations can overlap in time.
We extract all such simultaneous action pairs for both training ($2851$ motions), and validation sets~($1232$ motions).
We only consider the sequences that have a length between 600~(20~sec.) and 15~(0.5~sec.) frames. From the validation set, we exclude redundant pairs with the label `stand', because this commonly occurs in the data while not
representing challenging cases.
We also remove pairs that are \textit{seen} in the training set,
and end up with  $667$ sequences that contain two simultaneous actions.
The results on the full validation set are provided in
\if\sepappendix1{Section~E.4}
\else{Section~\ref{subsec:supmat:full_test_set}}
\fi
of the Appendix.
Besides the simultaneous pairs, 
we include the single-action data from BABEL in training. Specifically,
there are $24066$
and $8711$ single-action motions for training and validation sets, respectively.
In our experiments, we denote the simultaneous actions from BABEL with \textbf{Real-Pairs}, the single-motion
segments from BABEL with \textbf{Real-Singles}, and our synthetic data created by using body-part labels from GPT with \textbf{Synth-Pairs}. We perform evaluation only on the real spatial pairs of the BABEL validation set
to assess the quality of simultaneous action generation. 
\cam{We use the validation set as test set and train all of our models for 500 epochs.}

We report evaluation metrics adopted by \cite{Ghosh_2021_ICCV,petrovich2022temos,TEACH:3DV:2022}: Average Positional Error (APE), and Average Variational Error (AVE). However,
we observe that these metrics do not always correlate well with the visual quality of motions,
nor their semantic correspondence. We introduce, and additionally report, a new
\textbf{TEMOS score}, which compares the cosine similarity between the generated motion
and the ground truth after encoding them into the motion encoder of TEMOS~\cite{petrovich2022temos}, which is trained on BABEL Real-Singles (we do not observe significant changes when altering this model with TEMOS trained on different data; see 
\if\sepappendix1{Section~E.2}
\else{Section~\ref{subsec:supmat:different_temos_scores}}
\fi
of the Appendix). 
This is similar in spirit to BERTScore~\cite{bertscore2020}, which evaluates text generation quality by comparing to the ground truth
in the text embedding space.
More details can be found in 
\if\sepappendix1{Section~D}
\else{Section~\ref{sec:supmat:temos_score}}
\fi
of the Appendix.
While this metric is also imperfect (e.g., it still assumes a single ground truth action), we observe that it better correlates with realism and motion semantics as it has been trained to encode motions controlled by text descriptions.
An alternative performance measure is adopted
by \cite{Guo2022GeneratingDA} that reports motion-to-text
retrieval metrics, randomly selecting for each motion
31 negative text descriptions
along with the ground truth.
\cam{Finally, we include diversity metrics in
	\if\sepappendix1{Section~E.3}
	\else{Section~\ref{subsec:supmat:diversity}}
	\fi
	of the Appendix.
}

\subsection{Single-action baselines}
\label{subsec:baselines}

In the following, we introduce and describe two baselines
using a model trained with one description per motion:
(i) A naive single-action baseline that relies on a text-to-motion synthesis
model trained on single actions, tested on pairs of actions.
(ii) Our proposed GPT-compositing applied on independent
motion generations from a single-action model.

\noindent\textbf{Single-action model.}
Our first baseline %
tests the ability of single-action models to synthesize compositions by only modifying the input text. We train with Real-Singles from BABEL. At test time, we concatenate the text descriptions using `while' as a keyword and evaluate the generated motions.

\noindent\textbf{Single-action GPT-compositing.}
Another single-action baseline  generates two independent motions given two texts,
which are then combined using our proposed GPT-guided composition, stitching body parts from two motions (as described in our synthetic data creation; see Section~\ref{subsec:synth}).
Note that unlike the synthetic data, which combines real motions, this baseline
combines generated motions.
The disadvantage of this model is that it requires GPT at test time, and is based on heuristics
that may be error-prone, such as trimming the motions to the same duration, and resolving common body part labels (see \cam{the} supplementary video \cam{on our project page} for details). %
In the presence of a model that is trained only on individual actions
(Real-Singles), we observe that the GPT-based compositing of two independent generations %
improves the performance over the single-action baseline (as shown in Table~\ref{tab:archi_design} top).
Based on qualitative observation (see Section~\ref{subsec:qualitative}),
the single-action baseline often generates one out of the two actions.
The GPT-compositing baseline %
better captures both actions; however, lacks realism due to composing actions with heuristics.
SINC, which trains on compositional data,
alleviates both issues.

\subsection{The effect of the input text format}
\label{subsec:ablations}

To confirm whether our free-form input format
sacrifices performance compared to a more controlled
alternative of keeping the two action texts separate,
we experiment with a variant of our SINC model
by changing the text encoding.
Instead of a single text combining two actions,
we concatenate them together with a learnable
separation token in between after independently
encoding the actions with DistilBERT.
We refer to this separate text encoding variant as SINC-STE.
In Table~\ref{tab:archi_design}, we compare SINC
with SINC-STE
when trained only with Real-Pairs,
and observe a better TEMOS score with the free-form
text augmentations, at the cost of worse positional errors.
We observe that metrics based on joint positions may score high even in the absence of the second action, especially if it involves a fine-grained motion
(see supplementary video).
Besides quantitative performance, SINC has the advantage
of allowing more flexible inputs.

\subsection{Training with different sets of data}
\label{subsec:experiments_data}

\noindent\textbf{Contribution of Real-Singles and Real-Pairs.}
In Table~\ref{tab:archi_design}, we report the performance of SINC when adding both Real-Pairs
and Real-Singles to training.
We see that
training with the large number of single actions of BABEL,
in addition to the small amount of action pairs, improves performance,
and highlights the limited scale of the available pairs.

\begin{figure*}
	\centering
	\includegraphics[width=\textwidth]{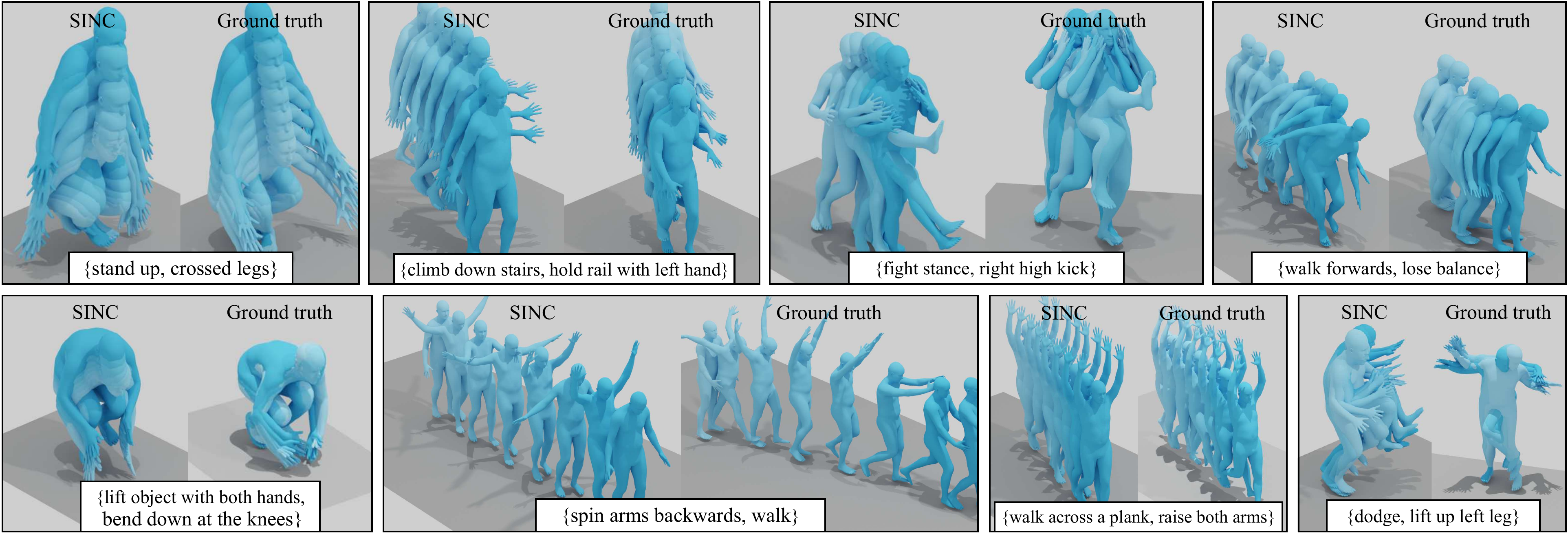}\\ 
	(a) \\ %
	\includegraphics[width=\textwidth]{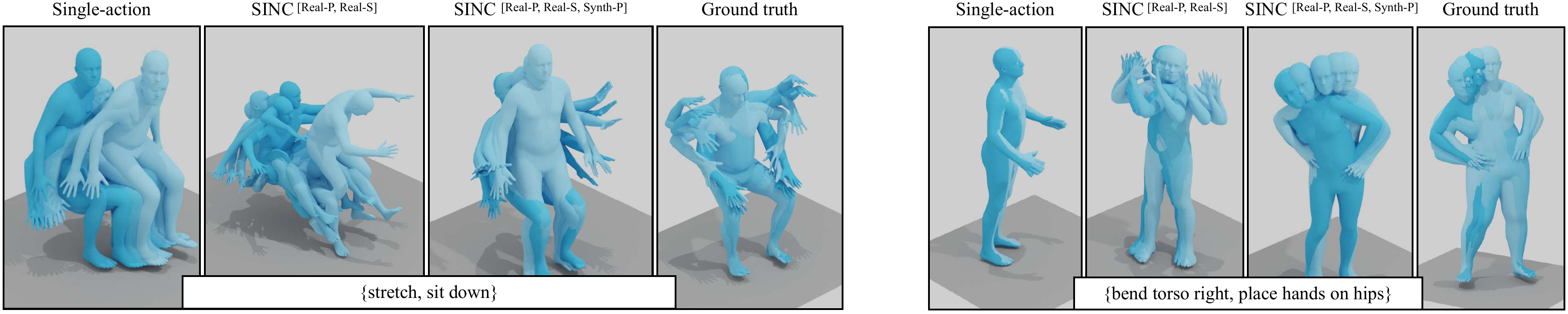}\\ 
	(b) \\ %
	\caption{\textbf{Qualitative analysis:}
		(a) We present qualitative results for our final model, \methodname, %
		for various description pairs
		from the validation set. 
		Our generations correctly correspond to the input semantics
		even when they are different from the ground truth,
		highlighting the challenge of coordinate-based (positional) performance measures.
		We display the ground truth (GT) for reference to define what the given actions mean.
		(b) We compare different models on two
		simultaneous action pairs. Both the Single-action model and the model not trained on synthetic data fail to generate those two compositions. Our model trained with the synthetic data successfully generates the composition in both cases. We include more %
		comparisons in \cam{the} supplementary video \cam{on our project page}.
	}
	\label{fig:qualitative}
\end{figure*}

\noindent\textbf{Contribution of GPT-guided Synth-Pairs.}
We experiment with different training sources in Table~\ref{tab:synthetic},
mainly to assess the effect of adding synthetic training data.
The percentages (0, 50, or 100) reflect the probability $p$
that a real-single action is composited synthetically with another action
(see Section~\ref{subsec:implementation}).
When using all training data (i.e., Real-P, Real-S 50\%, Synth-P 50\%),
we obtain the best TEMOS score, and more importantly
observe better qualitative results (see Figure~\ref{fig:qualitative}).
In particular, the model trained with GPT-guided synthetic
data demonstrates superior generalization capability to unseen
combinations.
In the supplementary video, we provide results
with input combinations that are unseen both in the real training
and validation sets.

\begin{figure}
    \centering
    \includegraphics[width=\columnwidth]{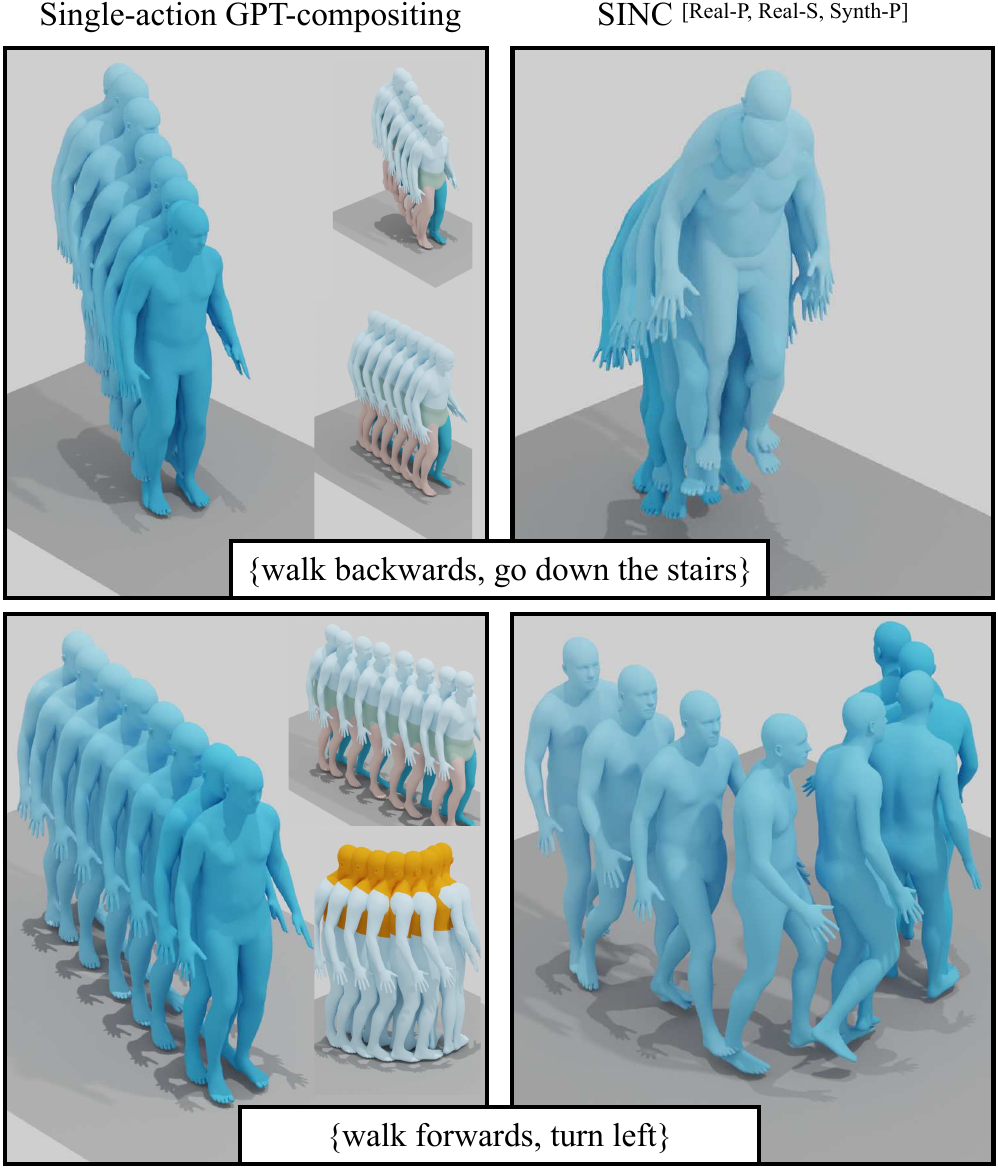}\\ 
        \caption{\textbf{Single-action GPT-compositing vs SINC:} We show two examples that highlight the advantage of our model compared to GPT compositions. Top: The detected body parts overlap causing the stitching to generate a forwards movement. Bottom: The global orientation is taken from the `walk forwards' failing to generate a left turn.
    }
    \label{fig:sinc-vs-gpt}
\end{figure}

\noindent\textbf{Synthetic data without GPT guidance.}
We further test whether 
our GPT-guidance to generate synthetic data is better than
just randomly mixing body parts (Random composition).
In Table~\ref{tab:synthetic},
GPT compositions outperform Random compositions, especially
when training only on synthetic data (0.539 vs 0.618 TEMOS score).

\subsection{Qualitative analysis}
\label{subsec:qualitative}

In Figure~\ref{fig:qualitative}~(a), we present simultaneous action generations using \methodname for the validation set of BABEL.
We show one random generation from our model for each
description pair (left), along with the ground truth (right).
Note that we display one sample due to space constraints,
but the model can synthesize
multiple diverse motions per input.
We observe that, while being sometimes different from the ground-truth
motion, our generations follow the semantics of
\textit{both} actions, achieving spatial compositionality.
Moreover, we qualitatively compare different models trained with and without synthetic data in Figure~\ref{fig:qualitative}~(b),
for the pair~ \{`stretch', `sit down'\} and~ \{`bend torso right', `put hands on hips'\}.
This action pair combination is unseen in Real-Pairs, but is seen 
in the Synthetic-Pairs data. In both cases, the Single-action model and  the model that has not been trained on Synthetic-Pairs~(first two columns) fail to generate the motion in contrast to \methodname which is trained  on spatial compositions.

Finally, in Figure \ref{fig:sinc-vs-gpt} we show failure cases of GPT-composition. Our baseline fails to generate a motion that corresponds to the instruction when the body parts are overlapping (top row). Another failure case happens when global orientation is important for the semantics of an action (`turn left') and is assigned to the walking action since it involves both feet (bottom row).

\subsection{Limitations}
\label{subsec:limitations}

Our framework relies on synthetic data
creation by combining arbitrary motions together.
Even if the body parts are compatible, in real life, not all
actions appear simultaneously together.
Future work should also explore the {\em semantic} compatibility between
actions by extracting this knowledge
from language models to construct semantically
meaningful compositions.
However, language models are also prone to mistakes.
In particular, GPT-3 body part labels
may be insufficient or ambiguous (e.g., `walking' may
or may not involve hands).
Additionally, going beyond our 6 course parts to obtain fine-grained body part label association is important.
In particular, this could involve the fingers and even facial expressions.
Another limitation of our work (and the whole field) concerns
the evaluation metrics. Despite
introducing a new TEMOS score,
perceptually meaningful  performance measures are still missing.
Finally, 
our model is conceptually not limited to pairs,
but since it is rare to simultaneously perform more than two actions,
we only focus on pairs in this work.

\section{Conclusions}
\label{sec:conclusions}

In this work, we established a new method to
create spatial compositions of 3D human motions.
Given a set of textual descriptions, our \methodname model
is able to generate motions that simultaneously
perform multiple actions presented as textual input.
We make use of the GPT-3 language model to obtain
a mapping between actions and body parts to automatically create synthetic combinations of compatible actions.
We use these synthetic motions to enrich the training of our model and find that it helps it generalize to new, complex, motions.
We introduce multiple baselines and experiment with different
data sources for this new problem. Our findings
will open up possibilities for further research
in fine-grained motion synthesis.
While here we focus on spatial composition, 
future work should explore
jointly modeling spatial and temporal action
composition.

\bigskip

{ %
\noindent\textbf{Acknowledgments.}
This work was granted access to
the HPC resources of IDRIS under the allocation 2022-AD011012129R2 made by GENCI.
GV 
acknowledges the ANR project CorVis ANR-21-CE23-0003-01.
The authors would like to thank Peter Kulits for proofreading and Benjamin Pellkofer for IT support.

\noindent\textbf{MJB Disclosure:} \url{https://files.is.tue.mpg.de/black/CoI_ICCV_2023.txt}
\par
}

{\small
\bibliographystyle{ieee_fullname}
\bibliography{references}
}

\bigskip
{\noindent \large \bf {APPENDIX}}\\

\renewcommand{\thefigure}{A.\arabic{figure}} %
\setcounter{figure}{0} 
\renewcommand{\thetable}{A.\arabic{table}}
\setcounter{table}{0} 

\appendix

This document provides
additional details about our method and experiments.
\cam{In particular, we evaluate our synthetic data approach on a recently proposed diffusion model~\cite{Chen2023}~(Section~\ref{sec:supmat:mld}),} 
elaborate on our GPT-based body-part annotation method (Section~\ref{sec:supmat:body_parts}),
our synthetic data creation pipeline (Section~\ref{sec:supmat:synth_data}), and our proposed TEMOS score (Section~\ref{sec:supmat:temos_score}).
We also provide additional quantitative evaluations 
(Section~\ref{sec:supmat:quantitative}).

\medskip
\noindent \textbf{Supplementary video.} Along with this document,
we provide a video, 
available on the \href{https://sinc.is.tue.mpg.de}{project page}, 
which includes visualizations of a sample of generated motions; these are difficult to convey in a static document.
(i) We first briefly describe our goal, motivation, and method. 
(ii) We then introduce baselines and illustrate their failure modes.
(iii) We provide qualitative comparisons against baselines, while highlighting limitations of the coordinate-based \textit{APE} metric.
(iv) Finally, we demonstrate the ability of our model to generalize to out-of-distribution input combinations, as well as combinations beyond pairs.
\section{\cam{Additional experiment with diffusion models}}
\label{sec:supmat:mld}
\cam{To complement our study with the TEMOS model~\cite{petrovich2022temos},
here, we provide an additional experiment by training
a more recent state-of-the-art architecture for text-conditioned motion generation.
Specifically, we implement Motion Latent Diffusion (MLD)~\cite{Chen2023}
with the same text input pipeline as our method~(see
\if\sepappendix1{Section~3.2).}
\else{Section~\ref{subsec:model}).}
\fi
Since MLD applies the diffusion on the latent space, we extract a single latent vector
per motion (using the TEMOS model trained on Real-singles as a feature extractor).
We train the diffusion model for 1000 epochs on 2 GPUs, with a batch size of 16, and learning rate of 1e-4.
Instead of the coordinate-based representation of Guo~et~al.~\cite{Guo2022GeneratingDA}, we directly
train on 6D rotation representation (as is done for TEMOS, see 
\if\sepappendix1{Section~3.3).}
\else{Section~\ref{subsec:implementation}).}
\fi
Apart from those adaptations, we use the same architectural choices as in the original paper~\cite{Chen2023}.
In Table~\ref{tab:supmat:mld}, we report the results
with and without synthetic data,
as we did for TEMOS in the main paper with
the rows 10 and 2 of
\if\sepappendix1{Table 3,}
\else{Table~\ref{tab:synthetic},}
\fi
respectively.
The same conclusion holds for MLD:
the model trained on additional synthetic data demonstrates better performance than the one
trained only on real data (Real-Pairs and Real-Singles).

\begin{table} %
    \centering
    \begin{NiceTabular}{lcc}
    \toprule
    Model & Synthetic training & TEMOS Score  \\
    \midrule
    MLD~\cite{Chen2023} & \xmark & 0.612 \\
    MLD~\cite{Chen2023} & \cmark & \textbf{0.638} \\
    \midrule
    TEMOS~\cite{petrovich2022temos} & \xmark &0.640 \\
    TEMOS~\cite{petrovich2022temos} & \cmark & \textbf{0.644} \\
    \bottomrule
    \end{NiceTabular}
    \caption{\cam{\textbf{Additional results with a diffusion model:} We report the performance of MLD~\cite{Chen2023}
    		with and without adding the synthetic training data. We observe that synthetic data helps
    		for both MLD and TEMOS.
    }
    }
    \label{tab:supmat:mld}
\end{table}

}

\section{Body Part Labeling with GPT-3}
\label{sec:supmat:body_parts}
BABEL includes $6518$ unique language labels for training and validation. We use these raw labels as input in the GPT-3 query.
We prompt the public API \url{https://openai.com/api/} for each of the BABEL action labels and %
automatically retrieve the body parts that are involved in the motion.
We experimented with various prompts before deciding on our final prompt template. We observed that GPT-3 
outputs are easier to parse and map to our predefined list of body parts if we provide this list, as well as few-shot examples consisting of question-answer pairs.
We use the following prompt, to extract the body part annotations for our synthetic data creation, as described in \if\sepappendix1{Section~3.1:}
\else{Section~\ref{subsec:synth}:}
\fi
\begin{lstlisting}[language={}, caption=\textbf{GPT prompt template}, label={lst:prompt}]
The instructions for this task are to choose 
your answers from the list below:

left arm
right arm
left leg
buttocks
waist
right leg
torso
neck

Here are some examples of the question and answer 
pairs for this task:

Question: What are the body parts involved in the
action of: walk forwards?
Answer: right leg
left leg
buttocks

Question: What are the body parts involved in the
action of: face to the left?
Answer: torso
neck

Question: What are the body parts involved in the
action of: put headphones over ears?
Answer: right arm
left arm
neck

Question: What are the body parts involved in the
action of: sit down?
Answer: right leg
left leg
buttocks
waist

Question: What are the body parts involved in the 
action of: [ACTION]?
\end{lstlisting}
Listing~\ref{lst:prompt} shows the full prompt used to extract the annotations using GPT-3 for composing actions spatially.
In
\if\sepappendix1{Table 1,}
\else{Table~\ref{tab:sgpt-labels}}
\fi
of the main paper,
we quantitatively evaluated the body part labeling performance of this prompt, along with alternative prompts. Here, in Table~\ref{tab:supmat:gpt-examples}, we provide qualitative examples to illustrate the behavior of GPT-3 to each of the prompt types.
(a) ``Free-form'' prompt type contains only L40-41 from Listing~\ref{lst:prompt}. (b) ``Choosing from a list'' contains both L1-11, L40-41. (c) ``Choosing from a list + Few-shot examples'' refers to the full prompt.
As shown in Table~\ref{tab:supmat:gpt-examples}, using ``Free-form'' prompting requires a tedious post-processing of GPT-3 responses, since one needs a comprehensive mapping from all possible body part namings to our list. Moreover, the level of details is not consistent across actions~(e.g., `left leg and hips' versus `deltoid and triceps muscles').
We extract the associated body parts by detecting keywords
from a manually constructed lookup table; %
however, the labeling accuracy based on 
\if\sepappendix1{Table 1}
\else{Table~\ref{tab:sgpt-labels}}
\fi
of the main paper is still lower than instructing GPT-3 to choose from a list. 
We obtain further gains by including few-shot examples in the prompt. 
This is demonstrated qualitatively in Table~\ref{tab:supmat:gpt-examples}
for the label `rotate shoulders' which GPT-3 includes neck in addition to torso or `walk backwards with arms attach to the waist' for which arms are mistakenly omitted for the ``Choose from a list'' prompt.
Our final prompt that provides both the list and few-shot examples perform best, while also requiring significantly less post-processing.

\begin{table*}
\centering
\resizebox{\linewidth}{!}{

\begin{tabular}{l  l  l} 
\toprule
\multicolumn{1}{c}{\textbf{Action}}  & \multicolumn{1}{c}{\textbf{Prompt Type}}  & \multicolumn{1}{c}{\textbf{GPT-3 Response}}   \\ 

\midrule

\multirow{3}{*}{move right arm in circular motion}                 & Free-form                                & The person's right arm, shoulder, and possibly the upper part of their body.                                                       \\ 
                                                                   & Choosing from a list                     & right arm                                                                                                                          \\ 
                                                                   & Choosing from a list + Few-shot examples & right arm                                                                                                                          \\ 
\midrule
\multirow{3}{*}{make large circles with left leg in front of body} & Free-form                                & The left leg and the hips                                                                                                          \\ 
                                                                   & Choosing from a list                     & left leg                                                                                                                           \\ 
                                                                   & Choosing from a list + Few-shot examples & left leg                                                                                                                           \\ 
\midrule

\multirow{3}{*}{overhead throw}                                    & Free-form                                & The deltoid muscle in the shoulder and the triceps muscle in the arm are moving when someone is doing an overhead throw.           \\ 
                                                                   & Choosing from a list                     & left arm right arm                                                                                                                 \\ 
                                                                   & Choosing from a list + Few-shot examples & left arm right arm torso                                                                                                           \\ 
\midrule
\multirow{3}{*}{walk backwards with arms attach to the waist}      & Free-form                                & The body parts involved in the action of walking backwards with arms attached to the waist are the legs, arms, back, and abdomen.  \\ 
                                                                   & Choosing from a list                     & right leg left leg buttocks                                                                                                        \\ 
                                                                   & Choosing from a list + Few-shot examples & left arm right arm left leg right leg waist                                                                                        \\ 
\midrule
\multirow{3}{*}{put down bottle with left hand}                    & Free-form                                & Left arm Left hand Fingers                                                                                                         \\ 
                                                                   & Choosing from a list                     & left arm                                                                                                                           \\ 
                                                                   & Choosing from a list + Few-shot examples & left arm torso                                                                                                                     \\ 
\midrule
\multirow{3}{*}{rotate shoulders}                                  & Free-form                                & The body parts involved in the action of rotating the shoulders are the neck, shoulders, arms, and back.                           \\ 
                                                                   & Choosing from a list                     & left arm right arm arm torso neck                                                                                                  \\ 
                                                                   & Choosing from a list + Few-shot examples & left arm right arm arm torso                                                                                                       \\
\bottomrule
\end{tabular}
}
\smallskip
    \caption{\textbf{GPT response examples for different prompt types:} 
    We show  the responses of GPT-3 on some examples that demonstrate the differences between different prompt types (see
    \if\sepappendix1{Table 1}
    \else{Table~\ref{tab:sgpt-labels}}
    \fi
    of the main paper).
    The output of the free-form prompt is non-trivial to parse and map to our list of body parts. On the other hand, providing the list and few-shot examples
    encourages GPT-3 to follow a more strict format, and to describe the body parts with the same words as in our list.
    }
    \label{tab:supmat:gpt-examples}
\end{table*}

We explain the reasoning behind replacing `global orientation' with `waist' and `buttocks' in the list of body parts.
In our initial prompts we used `global orientation' as part of the list.
However, we observed that the model frequently returned `waist' and `buttocks' even when they were not in the list.
Furthermore, GPT-3 responses included `global orientation' even in cases when it was not necessary e.g., `lift arm', `raise leg'. Consequently, we chose to remove `global orientation', and add `waist' and `buttocks' instead.

Finally, we include the label `neck' in addition to `torso`, since GPT-3 tends to include `neck' in its responses, especially when we prompt for the actions: `look left' / `look right'.

\begin{figure}
    \centering
    \includegraphics[trim=100 100 60 80,clip, width=0.8\linewidth]{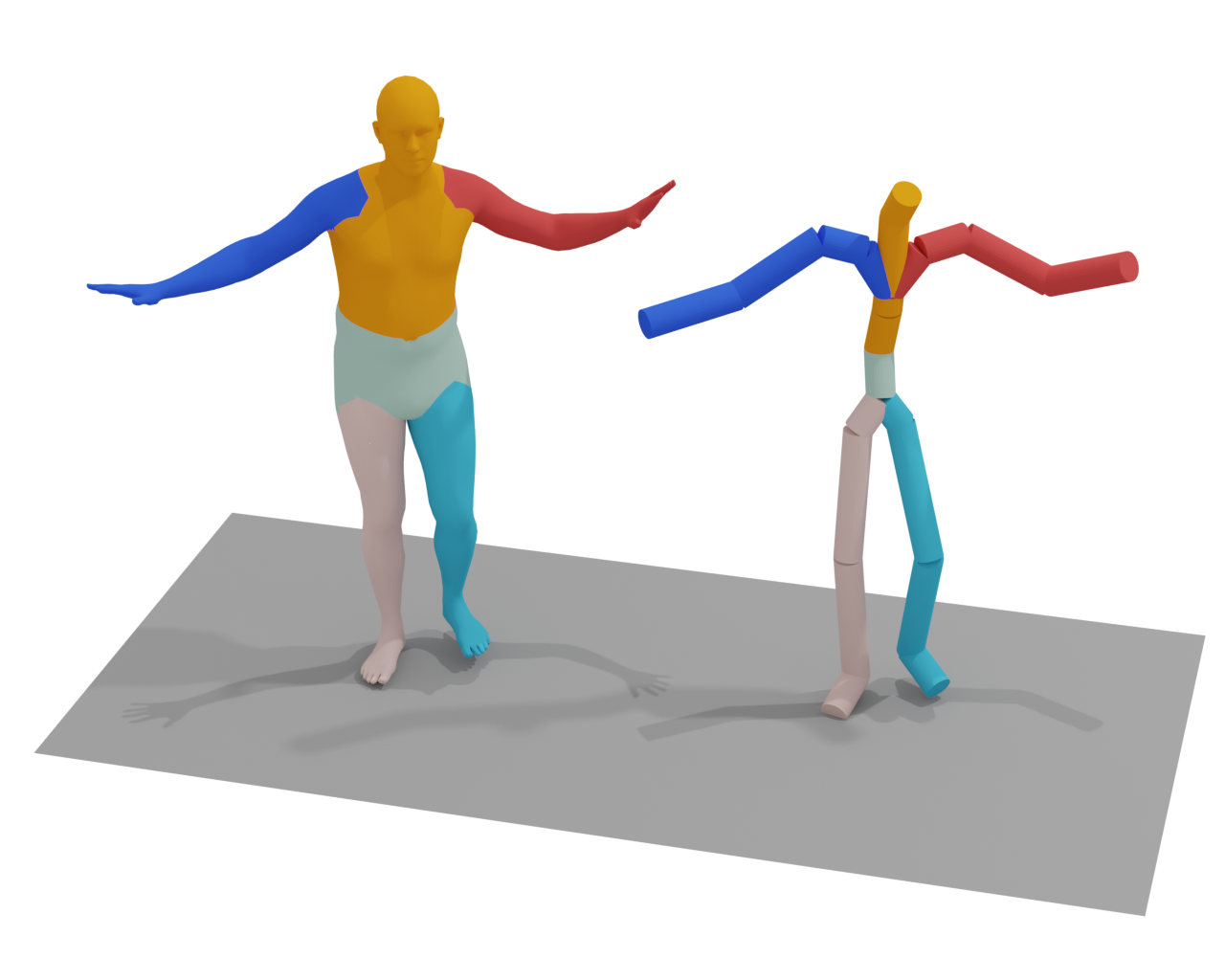}
    \caption{\textbf{Body parts:} Each color indicates a different body part.
    Vertices (left) and the skeleton (right) are extracted from the SMPL body model.
    }
    \label{fig:supmat:body_parts}
\end{figure}

We use $6$ body part labels based on common body segmentation maps~\cite{SMPL:2015}, which we show in Figure~\ref{fig:supmat:body_parts}. Since most of the AMASS dataset does not contain fine-grained hand motions, we do not include hands as a separate body part.
Furthermore, we observe that GPT-3 behavior may become inconsistent %
if we provide a long list of fine-grained parts instead of
few coarse labels.
The main body parts include the right/left extremities, the torso-neck part and the pelvis denoted as global.
Global, except for the pelvis rotation, includes global translation of the body and it is used when either the waist or buttocks are included in GPT-3's response.

\begin{figure} %
    \centering
    \includegraphics[width=0.5\linewidth]{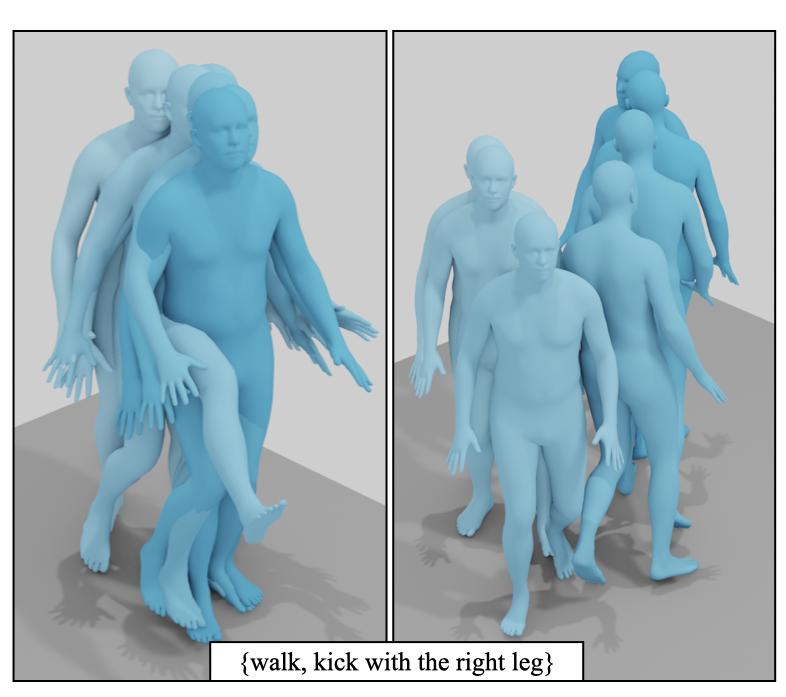}
    \caption{\cam{\textbf{Testing incompatible simultaneous actions:} 
    		We apply our model SINC on the input \{`walk', `kick with the right leg'\}
    		which represents an example of two incompatible actions due to involving
    		the same body part `right leg'. We display
    		two random generations from our model, once with 2-second duration (left), and once with 4 seconds (right). We observe that
    		SINC generates one of the two actions in each sample (`kick' on the left, `walk' on the right).
    	}
    	}
    \label{fig:supmat:ambiguity}
\end{figure}

\section{Synthetic Data Creation}
\label{sec:supmat:synth_data}
\begin{figure}
    \centering
    \includegraphics[width=0.9\linewidth]{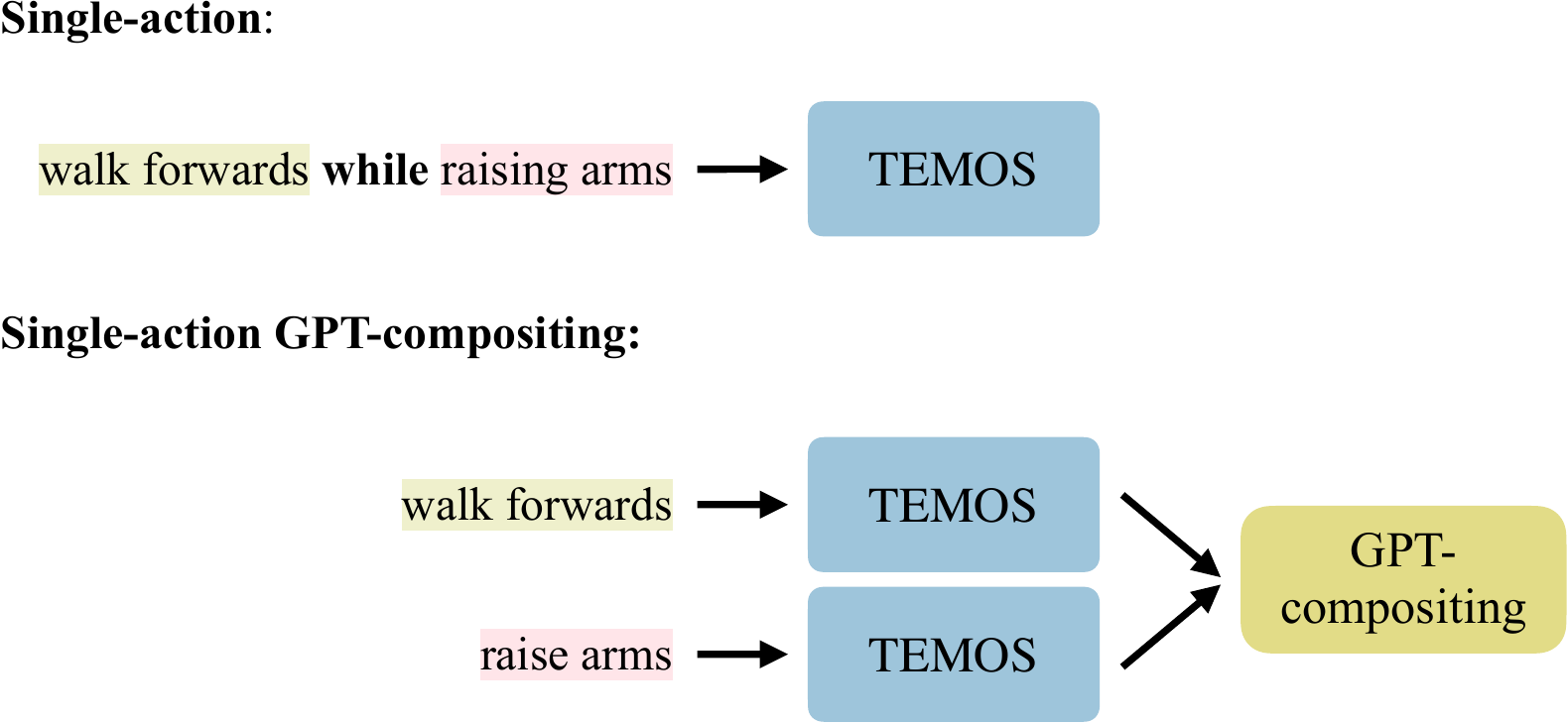}
    \caption{\textbf{Single-action baselines:} For both baselines, TEMOS is trained on Real-Singles of BABEL. On the top, we concatenate the textual inputs by adding the word ``while'' in between actions. On the bottom, we generate the two actions independently and combine them with the body part guidance from GPT-3.
    }
    \label{fig:supmat:baselines}
\end{figure}

We use GPT-3-guided spatial compositions in two parts of this work. First, we use GPT-3 to benchmark how well a single-action baseline can perform,
by applying composition as post-processing 
on independently generated motions (Figure~\ref{fig:supmat:baselines} bottom). %
Secondly, we use GPT-3 to create synthetic data to train %
our model.
In both cases, we employ the method described in \if\sepappendix1{Section~3.1}
\else{Section~\ref{subsec:synth}}
\fi
of the main paper. 
We use the heuristic of stitching the motion 
with less body parts (motion B) on top of the other motion (motion A), 
because the body parts of motion B are more likely to be
local (as in ``waving the right hand'') 
and important for keeping the semantic of the motion. 
On the other hand, motion A is more likely 
to be a global motion (as in ``walking'' or ``sitting'') and grafting motion B onto motion A usually produces a realistic motion and preserves the semantics of both motions. Note that these heuristics were determined based on visual inspection over several examples, and may not be optimal.

The difference in the case of synthetic data creation is the compatibility test, which makes sure that no body part is involved in both of the motions being composited. Moreover, synthetic data combines existing real motions, and the single-action baseline combines generated motions.

\cam{We only apply the compatibility check for the synthetic data
	generation to avoid composing invalid motions, since a human
	can physically not perform two actions with the same body part in most cases.
	This choice was simply to ensure better synthetic data quality, as without it,
	the composition may be reduced down to one action (e.g., ‘walking’ would overwrite ‘kicking’
	as the leg cannot do both). At test time, when we query ‘walk’ and ‘kick with the right leg’ with two different durations,
	SINC randomly generates one of the two actions, as seen in Figure~\ref{fig:supmat:ambiguity}.}

\begin{table*}[t]
\centering
\resizebox{\linewidth}{!}{
\begin{NiceTabular}{lll|c|rrrr|rrrr}
        \toprule
          \multirow{2}{*}{\textbf{Conjuction Word}} & \cam{\textbf{Seen during}} & \multirow{2}{*}{\textbf{Model}} & TEMOS $\uparrow$ &
         \multicolumn{4}{c}{Average Positional Error $\downarrow$} & \multicolumn{4}{c}{Average Variance Error $\downarrow$} \\
          & \cam{\textbf{training}} & & score & \small{root joint} & 
         \small{global traj.} & \small{mean local} & \small{mean global} &
         \small{root joint} & \small{global traj.} & \small{mean local} & \small{mean global}  \\
        \midrule
      \multirow{2}{*}{\texttt{while}} & \multirow{2}{*}{\cam{\cmark}} & Single-action  & 0.601 & 0.592 & 0.551 & 0.286 & 0.712 & 0.076 & 0.075 & 0.013 & 0.083  \\
       & & \textbf{SINC}  & \textbf{0.644} & 0.493 & 0.463 & 0.266 & 0.616 & 0.066 & 0.065 & 0.012 & 0.072 \\
       \midrule

        \multirow{2}{*}{\texttt{during}} & \multirow{2}{*}{\cam{\cmark}} & Single-action  & 0.598 & 0.629 & 0.587 & 0.284 & 0.752 & 0.085 & 0.084 & 0.013 & 0.093 \\
       & & SINC & \textbf{0.642} & 0.497 & 0.471 & 0.261 & 0.622 & 0.065 & 0.063 & 0.012 & 0.071 \\ 
       
       \midrule
       \multirow{2}{*}{\texttt{and ... at the same time}} & \multirow{2}{*}{\cam{\cmark}} & Single-action  & 0.599  & 0.607 & 0.568 & 0.283 & 0.722 & 0.084 & 0.083 & 0.014 & 0.092 \\
       & & SINC & \textbf{0.643} & 0.495 & 0.468 & 0.264 & 0.620 & 0.065 & 0.064 & 0.012 & 0.072 \\
    \midrule
    \midrule
    \multirow{2}{*}{\cam{\texttt{in parallel}}} & \multirow{2}{*}{\cam{\xmark}} & Single-action & 0.600 & 0.611 & 0.570 & 0.294 & 0.736 & 0.081 & 0.081 & 0.012 & 0.089 \\
       & & SINC & \textbf{0.643} & 0.583 & 0.555 & 0.266 & 0.704 & 0.074 & 0.072 & 0.012 & 0.080 \\
    \midrule
    \multirow{2}{*}{\cam{\texttt{whilst}}} & \multirow{2}{*}{\cam{\xmark}} & Single-action  & 0.599  & 0.551 & 0.511 & 0.288 & 0.670 & 0.073 & 0.072 & 0.012 & 0.080 \\
       & & SINC & \textbf{0.644} & 0.491 & 0.461 & 0.262 & 0.614 & 0.066 & 0.065 & 0.012 & 0.072 \\
    \midrule
    \multirow{2}{*}{\cam{\texttt{synchronously}}} & \multirow{2}{*}{\cam{\xmark}} & Single-action & 0.596 & 0.520 & 0.476 & 0.294 & 0.644 & 0.074 & 0.072 & 0.013 & 0.081  \\
       & & SINC & \textbf{0.637} & 0.520 & 0.492 & 0.261 & 0.644 & 0.0644 & 0.0632 & 0.011 & 0.070 \\
    \bottomrule
    \end{NiceTabular}
   }
   \smallskip
    \caption{\textbf{Evaluation using different conjunction words:}
    \cam{In 
    	\if\sepappendix1{Table 2}
    	\else{Table~\ref{tab:archi_design}}
    	\fi
    	of the main paper,
    	we evaluated the models with the conjunction word \texttt{while}. Here, we report performance when joining the two actions using other conjunction words, for both seen and unseen conjunction words during training.}
    We observe similar trends for the TEMOS scores and the positional metrics as for using \texttt{while} to join the actions. 
	Overall, performance of Single-action methods remains significantly inferior, especially for the TEMOS score.
	Note that SINC refers to our best model which is trained on both Real Singles, Real Pairs and Synthetic Pairs.}
    \label{tab:supmat:during}
\end{table*}

\begin{figure} %
    \centering
    \includegraphics[width=0.60\columnwidth, scale=0.5]{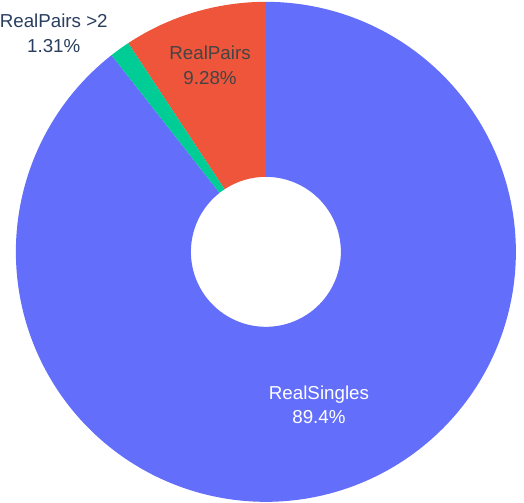}
    \caption{\textbf{Distribution of the training set:} The simultaneous Real-Pairs are the vast minority of the data, highlighting the importance of automatically enriching training data through our synthetic spatial compositions.
    }
    \label{fig:supmat:data-distribution}
\end{figure}

\section{TEMOS Score}
\label{sec:supmat:temos_score}

The position-based metrics typically used in prior work~\cite{Ghosh_2021_ICCV,petrovich2022temos,TEACH:3DV:2022} compare generated motions with the ground-truth motion in the coordinate space local to the body: they measure differences of positions and do not take into account semantics. %
Here are four types of examples where the metrics can fail: 1) with a cyclic motion such as ``walking'', the generation can be out of phase with the ground truth and still be semantically valid; 2) even for a non-cyclic motion such as ``throwing an object'', the timing can be different and can lead to bad scores on common metrics; 3) if the input text description is ambiguous such as ``kick'' (where the motion can be done from one leg or the other), the metrics may not reflect the quality of the generated motion; 4) if the motion demonstrates severe foot sliding or body translation artifacts, the error may be dominated by the translation error, effectively ignoring the overall implausibility of the limb motion e.g., feet not moving.

To avoid these issues, we introduce another performance measure called \textit{TEMOS score}. We train a TEMOS model on BABEL Real-Singles for 1000 epochs, freeze its weights, and use its motion encoder component.
Then, we extract features by feeding a motion $B$ to the motion encoder, and use the mean of the distribution as the feature vector $f$. This feature captures the semantics of the motion as the motion space has been trained to explicitly model motion-text matching, i.e., cross-modal embedding space.

To calculate the TEMOS score, we feed the ground truth and the generated motions to the motion encoder, and extract the feature vectors $f_{GT}$ and $f_{motion}$, respectively. Then we compute the score based on their cosine similarity as follows:
\begin{align*}
    \text{TEMOS score}(f_{GT}, f_{motion}) &= \frac{1}{2} \left( 1 + \frac{f_{GT} \cdot f_{motion}}{||f_{GT}|| \cdot\ ||f_{motion}||} \right).
\end{align*}
The range of this score is between $0$ and $1$, with a maximum at $1$, which occurs when the two motions are identical.

\begin{table}[tb] %
\centering
    \begin{tabular}{l rr}
        \toprule
        &  
        \multicolumn{2}{c}{Model used for TEMOS score} \\
        & \textbf{Single-action} & \textbf{SINC}  \\
        \midrule
        \textbf{Single-action} & 0.601 & 0.594  \\
        \textbf{SINC} & 0.644 & 0.637 \\ 
        \bottomrule 
    \end{tabular}
\smallskip
    \caption{\textbf{TEMOS score with various TEMOS models:} 
    We report performance using different trained models to compute the TEMOS score. While the absolute score slightly differs when measured with 
    a different model (e.g., 0.644 vs 0.637), the relative ranking of the models
    we compare remains the same.
    }
    \label{tab:supmat:temosscore}
\end{table}

\begin{table} %
    \centering
    \begin{NiceTabular}{lcc}
    \toprule
     & Div. $\rightarrow$ & Multimod. $\uparrow$ \\
    \midrule
    SINC  & 1.10 & 1.13 \\
    Real             & 1.34 & - \\
    \bottomrule
    \end{NiceTabular}
    \caption{\cam{\textbf{Diversity evaluation:} We report the diversity and multimodality metrics of \cite{Guo2022GeneratingDA} for our SINC model.}}
    \label{tab:supmat:diversity}
\end{table}

\begin{table*}[htbp!]
    \centering
    \resizebox{\linewidth}{!}{
    \begin{NiceTabular}{l|cc|c|rrrr|rrrr}
        \toprule
        \multirow{2}{*}{\textbf{Model}} & \multicolumn{2}{c}{Tr.~Data} & TEMOS $\uparrow$ & \multicolumn{4}{c|}{Average Positional Error $\downarrow$} & \multicolumn{4}{c}{Average Variance Error $\downarrow$} \\
         & Real-P & Real-S & score & \small{root joint} &  \small{global traj.}  & \small{mean local} & \small{mean global} & \small{root joint} & \small{global traj.} & \small{mean local} & \small{mean global}  \\
        \midrule 
        \textbf{Single-action} & \xmark  & \cmark & 0.607 & 0.516 & 0.483 & 0.262 & 0.626 & 0.067 & 0.066 & 0.012 & 0.073 \\
        
        \textbf{Single-action GPT-compositing}  & \xmark  & \cmark & 0.626 & 0.458 & 0.431 & 0.244 & 0.569 & 0.068 & 0.067 & 0.011 & 0.074 \\
        \midrule
        \midrule
        \textbf{SINC-STE} & \cmark & \xmark  &  0.630  & 0.502 & 0.477 & 0.249 & 0.616 & 0.074 & 0.074 & 0.010 & 0.08 \\
        \textbf{SINC} & \cmark & \xmark  &  0.634 & 0.602 & 0.586 & 0.243 & 0.704 & 0.084 & 0.083 & 0.011 & 0.091 \\
        \textbf{SINC}  & \cmark & \cmark &  \textbf{0.645}  & 0.519 & 0.495 & 0.248 & 0.632 & 0.078 & 0.077 & 0.010 & 0.084 \\
        \bottomrule 
    \end{NiceTabular}
    }
    \smallskip
    \caption{\textbf{Baseline comparison on the full validation set of BABEL:} 
    We observe similar trends with the filtered validation set reported in 
    \if\sepappendix1{Table 2}
	\else{Table~\ref{tab:archi_design}}
	\fi
	of the main paper.
    }
    \label{tab:supmat:archi_design_full}
\end{table*}

\begin{table*}[htbp!]
\resizebox{\linewidth}{!}{
\begin{NiceTabular}{l|ccc|c|rrrr|rrrr}
        \toprule
          \multirow{2}{*}{\textbf{Synthetic data}} & \multicolumn{3}{c}{Training Data} &
          TEMOS $\uparrow$ &
         \multicolumn{4}{c}{Average Positional Error $\downarrow$} & \multicolumn{4}{c}{Average Variance Error $\downarrow$} \\
          & Real-P & Real-S\,\% & Synth-P\,\% & score & \small{root joint} & 
         \small{global traj.} & \small{mean local} & \small{mean global} &
         \small{root joint} & \small{global traj.} & \small{mean local} & \small{mean global}  \\
        \midrule

  \textbf{N/A}  & \cmark & 0 &  0  &  0.634 & 0.602 & 0.586 & 0.243 & 0.704 & 0.084 & 0.083 & 0.011 & 0.091 \\
                & \cmark & 100 & 0 & 0.645  & 0.519 & 0.495 & 0.248 & 0.632 & 0.078 & 0.077 & 0.010 & 0.084 \\
  \midrule
         \textbf{Random composition} & \xmark & 50 & 50 & 0.551 & 0.575 & 0.534 & 0.259 & 0.664 & 0.072 & 0.071 & 0.011 & 0.078 \\
                                     & \xmark & 0 & 100 &  0.552 & 0.454 & 0.411 & 0.263 & 0.551 & 0.068 & 0.067 & 0.011 & 0.074 \\
                                     & \cmark & 50 & 50 & 0.619 & 0.396 & 0.362 & 0.242 & 0.504 & 0.060 & 0.059 & 0.010 & 0.067 \\
                                     & \cmark & 0 & 100 & 0.619  & 0.422 & 0.390 & 0.241 & 0.530 & 0.062 & 0.061 & 0.010 & 0.068 \\
    
    \midrule

   \textbf{GPT composition} & \xmark & 50 & 50 & 0.554  & 0.641 & 0.604 & 0.262 & 0.731 & 0.074 & 0.073 & 0.011 & 0.081 \\
        
                            & \xmark & 0 & 100 & 0.632 & 0.424 & 0.405 & 0.237 & 0.543 & 0.055 & 0.054 & 0.011 & 0.062 \\
 
                                & \cmark & 50 & 50 & \textbf{0.651} & 0.418 & 0.397 & 0.234 & 0.533 & 0.055 & 0.054 & 0.010 & 0.062 \\ 
                    
                                & \cmark & 0 & 100 & \textbf{0.645} & 0.472 & 0.453 & 0.237 & 0.581 & 0.053 & 0.053 & 0.010 & 0.060 \\
    \bottomrule 
     \end{NiceTabular}
   }
   \smallskip
    \caption{\textbf{Contribution of the synthetic data on the full validation set of BABEL:} 
    	We complement
    	\if\sepappendix1{Table 3}
    	\else{Table~\ref{tab:synthetic}}
    	\fi
    	of the main paper, by reporting on the full validation set (without any filtering).
    }
    \label{tab:supmat:synthetic_full}
\end{table*}

\section{Additional Quantitative Evaluation}
\label{sec:supmat:quantitative}

We report quantitative results
when evaluating with various conjunction words
(Section~\ref{subsec:supmat:conjunction}),
when using various TEMOS models to compute the TEMOS score
(Section~\ref{subsec:supmat:different_temos_scores}), 
\cam{when evaluating the diversity and multimodality metrics (Section~\ref{subsec:supmat:diversity})},
and, 
when evaluating on the full validation set for completeness
(Section~\ref{subsec:supmat:full_test_set}).

\subsection{More conjunction words} %
\label{subsec:supmat:conjunction}

\cam{In our main paper experiments, we used \texttt{while} as our conjunction word.}
For completeness, in Table~\ref{tab:supmat:during} we evaluate the Single-action method and \cam{our best} model
with \cam{other} conjunction
words at test time. %
We observe that the differences are minimal and the methods perform similarly across different conjunctions.
\cam{This is true for all conjunctions both seen and unseen during training. The performance is similar, likely due to the text embeddings mapping the expressions to similar points.}

\subsection{TEMOS score with various TEMOS models}
\label{subsec:supmat:different_temos_scores}
As mentioned in
\if\sepappendix1{Section~4.1}
\else{Section~\ref{subsec:data}}
\fi
of the main paper,
to report the TEMOS
score, we use a TEMOS model trained on Real-Singles of BABEL.
Here, we analyze whether the choice of the TEMOS model
has a large impact on the results when trained
on pairs. In Table \ref{tab:supmat:temosscore}, we observe that the TEMOS score trend is similar when computed with TEMOS models trained on Real-Singles (Single-action) or on all real and synthetic data (SINC).

\subsection{Diversity}
\label{subsec:supmat:diversity}
\cam{Following Guo et al.~\cite{Guo2022GeneratingDA}, 
we report the overall diversity (for all action pairs), and 
multimodality (i.e., per-action-pair diversity) in
Table~\ref{tab:supmat:diversity}. We measure the L2 distance
between the TEMOS embeddings of two sets of generations. 
For multimodality we sample 20 generations per description, and for diversity we generate 5 samples per description. 
Both metrics are computed for 300 random descriptions from the BABEL validation set. 
Real motions do not contain a sufficient number of motions for each 
action pair, thus the reason for omitting their multimodality.}

\subsection{Full validation set}
\label{subsec:supmat:full_test_set}
As explained in \if\sepappendix1{Section~4.1}
\else{Section~\ref{subsec:data}}
\fi
of the main paper,
we report all the results on a challenging subset of the validation set (i.e., without the action `stand', and using only unseen examples). Here, we provide the results on the full validation set for completeness. In particular, we repeat the \if\sepappendix1{Tables 2 and 3}
\else{Tables \ref{tab:archi_design} and \ref{tab:synthetic}}
\fi
of the main paper,
in Tables \ref{tab:supmat:archi_design_full} and \ref{tab:supmat:synthetic_full}. As expected, we observe slightly improved results overall on this `easier' validation set and the conclusions remain similar to the comparison in the main paper.

\end{document}